\documentclass[letterpaper, 10 pt, journal, twoside]{IEEEtran}

\IEEEoverridecommandlockouts                              

\usepackage[bookmarks=true]{hyperref}
\usepackage{graphicx}

\usepackage{caption}
\usepackage{subcaption}
\usepackage{physics, amsmath}
\usepackage{xcolor}
\usepackage{graphicx}

\usepackage{booktabs}
\usepackage{multirow}
\newcommand\at[2]{\left.#1\right|_{#2}}

\begin{document}
\begin{minipage}{17cm}
    \copyright 2025 IEEE.  Personal use of this material is permitted.  Permission from IEEE must be obtained for all other uses, in any current or future media, including reprinting/republishing this material for advertising or promotional purposes, creating new collective works, for resale or redistribution to servers or lists, or reuse of any copyrighted component of this work in other works
    \newline \newline
    You can cite this work as, \newline
    \texttt{
    @article\{sriganesh24bayesianstaircase, \newline
      author=\{Sriganesh, Prasanna and Shirose, Burhanuddin and Travers, Matthew\}, \newline
      journal=\{IEEE Robotics and Automation Letters\}, \newline
      title=\{A Bayesian Modeling Framework for Estimation and Ground Segmentation of Cluttered Staircases\}, \newline
      year=\{2025\}, \newline
      volume=\{10\}, \newline
      number=\{5\},\newline
      pages=\{4164-4171\},\newline
      doi=\{10.1109/LRA.2025.3549662\} \newline\}
    }
\end{minipage}

\title{
A Bayesian Modeling Framework for Estimation and Ground Segmentation of Cluttered Staircases}
\author{Prasanna Sriganesh$^{1}$, 
Burhanuddin Shirose$^{1}$, 
 and  
Matthew Travers$^{1}$
\thanks{Manuscript received: December, 13, 2024; Accepted February, 22, 2025. This paper was recommended for publication by Editor Giuseppe Loianno upon evaluation of the Associate Editor and Reviewers' comments.}
\thanks{ $^{1}$All authors are from The Robotics Institute, Carnegie Mellon University, USA. \scriptsize\texttt{\{pkettava, bshirose, mtravers\}@andrew.cmu.edu}} 
\thanks{Supplementary Video - \scriptsize\texttt{\url{https://youtu.be/8baHgQ_rGLs}}}
\thanks{Digital Object Identifier (DOI): see top of this page.}
}

\maketitle

\begin{abstract}
Autonomous robot navigation in complex environments requires robust perception as well as high-level scene understanding due to perceptual challenges, such as occlusions, and uncertainty introduced by robot movement.  For example, a robot climbing a cluttered staircase can misinterpret clutter as a step, misrepresenting the state and compromising safety. This requires robust state estimation methods capable of inferring the underlying structure of the environment even from incomplete sensor data. In this paper, we introduce a novel method for robust state estimation of staircases. To address the challenge of perceiving occluded staircases extending beyond the robot's field-of-view, our approach combines an infinite-width staircase representation with a finite endpoint state to capture the overall staircase structure. This representation is integrated into a Bayesian inference framework to fuse noisy measurements enabling accurate estimation of staircase location even with partial observations and occlusions. Additionally, we present a segmentation algorithm that works in conjunction with the staircase estimation pipeline to accurately identify clutter-free regions on a staircase. Our method is extensively evaluated on real robots across diverse staircases, demonstrating significant improvements in estimation accuracy and segmentation performance compared to baseline approaches. 

\end{abstract}
\IEEEpeerreviewmaketitle

\section{Introduction}
\IEEEPARstart{S}{taircases}, an ubiquitous feature of human-built environments throughout history, have enabled access to different levels within structures. With the increasing availability of legged robot platforms, significant strides have been made in enabling agile locomotion across diverse terrain, including staircases. However, ensuring safe navigation in complex real-world scenarios, such as cluttered or damaged staircases, remains a formidable challenge. This necessitates robots that can not only perceive their environment but also continuously update their knowledge about the environment by inferring the underlying structure.

Accurately perceiving and modeling staircases in such scenarios requires overcoming three key challenges: occlusions, limited field-of-view and sensor noise. Occlusions from objects or clutter can hide parts of the staircase, leading to incomplete estimate of staircase location. Moreover, a robot's limited field-of-view often restricts perception to a few steps at a time, hindering estimation of the staircase's overall location and orientation. Robot movement introduces sensor noise, causing errors in measurements of step height, depth, and curvature. These challenges necessitate a robust approach that integrates noisy sensor measurements with prior knowledge for accurate staircase estimation.

\begin{figure}[t!]
    \centering
    \includegraphics[width=0.85\linewidth]{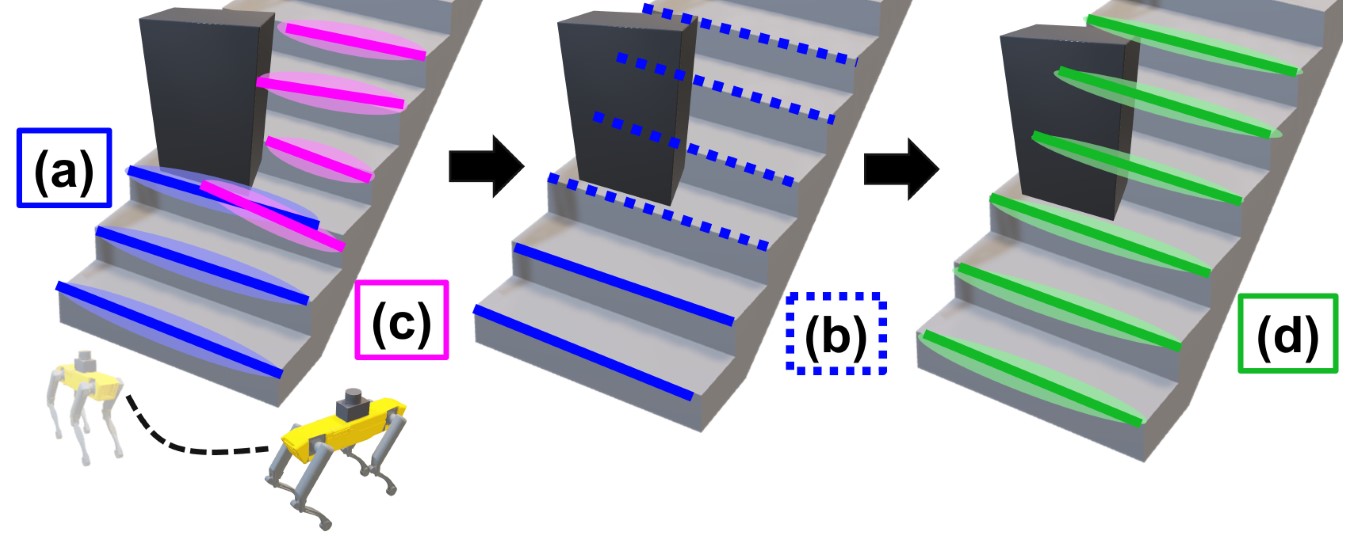}
    \caption{(a) Prior staircase estimate, $bel(\boldsymbol{_LX}_k)$ at time~$k$ (solid blue). (b)~Predicted staircase location, $p(\boldsymbol{_L\hat{X}}_{k+1|k})$ based on prior (dotted blue). (c)~New noisy measurement, $p(\boldsymbol{_LZ}_{k + 1})$ at time~$k\!+\!1$ (magenta). (d) Filtered staircase estimate $bel(\boldsymbol{_LX}_{k + 1})$ at time~$k\!+\!1$ by combing prediction and measurement (green).}
    \label{fig:intuition}
\end{figure}

\begin{figure}[t!]
    \centering
    \vspace{-1.0em}
    \begin{subfigure}[t]{0.27\linewidth}
        \centering
        \includegraphics[width = \linewidth]{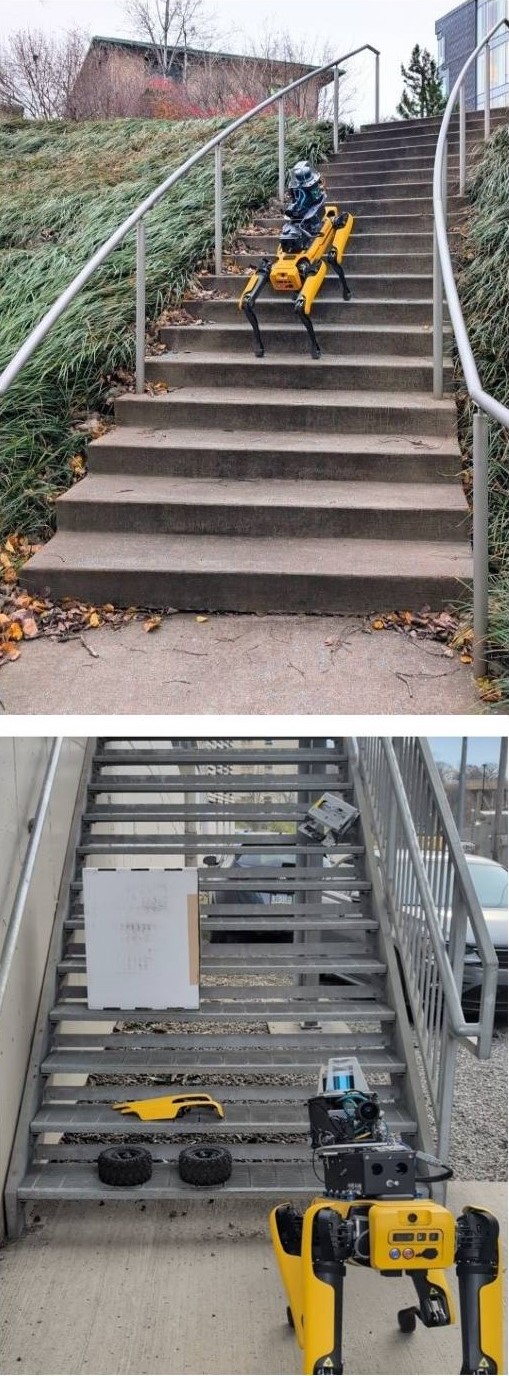}
        \caption{}
        \label{fig:opening_a}
    \end{subfigure}
    \begin{subfigure}[t]{0.29\linewidth}
        \centering
        \includegraphics[width = \linewidth]{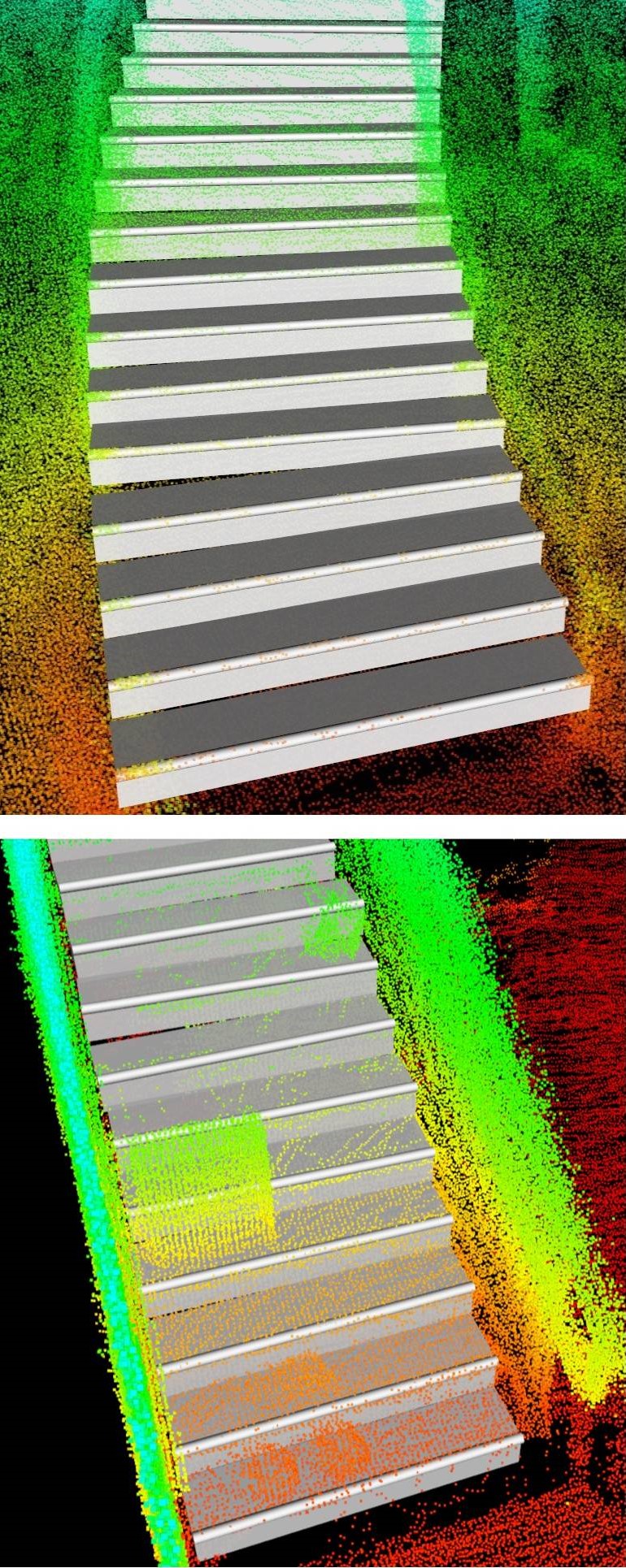}
        \caption{}
         \label{fig:opening_b}
    \end{subfigure}
    \begin{subfigure}[t]{0.29 \linewidth}
        \centering
        \includegraphics[width =\linewidth]{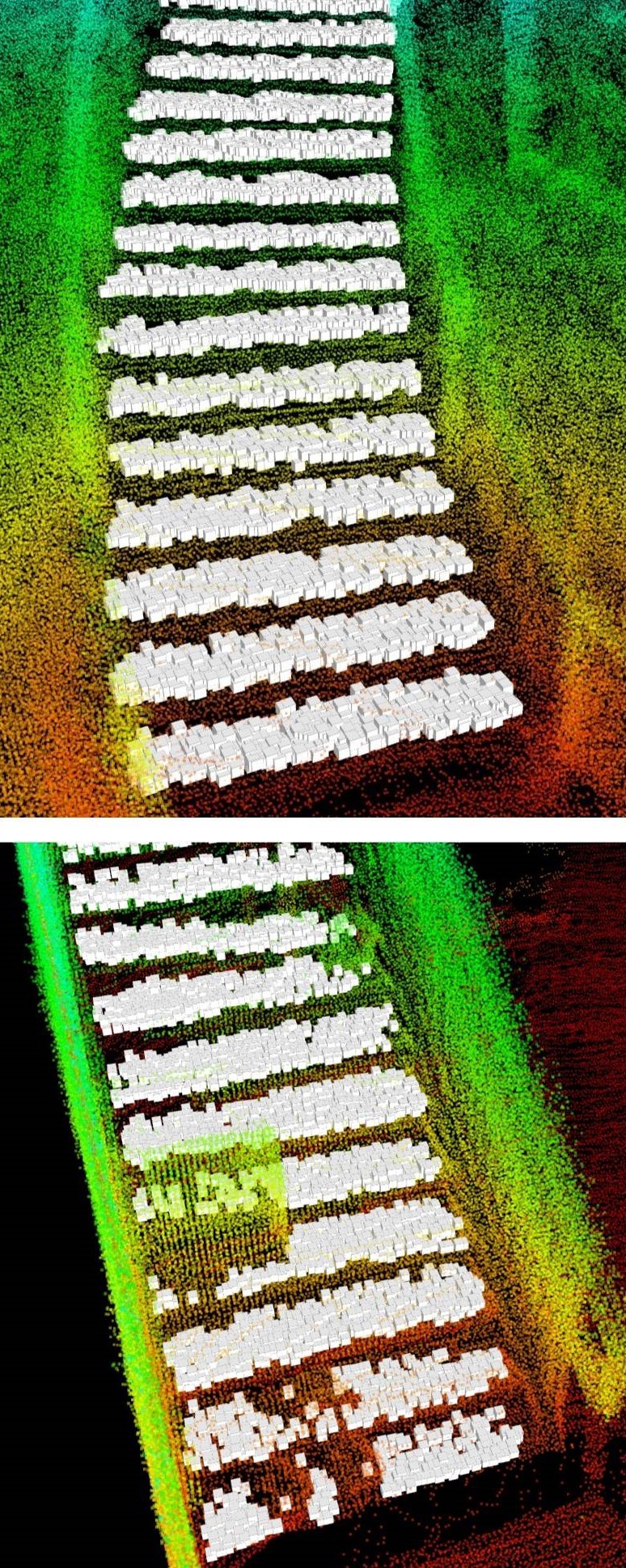}
        \caption{}
         \label{fig:opening_c}
    \end{subfigure}
    \caption{Results of our proposed staircase modeling and estimation framework. (a) Robot navigating staircase environments (b) Estimated staircase highlighted by white marker. (c) Segmented clutter-free regions on each stair.}
    \label{fig:opening_results}
    \vspace{-2.00em}
\end{figure}

In this paper, we propose a novel method for robust staircase estimation employing Bayesian inference, as illustrated in Fig. \ref{fig:intuition}. Our approach leverages prior knowledge about typical staircase geometry obtained from previously observed parts of the staircase to infer the location and geometry of new steps. This prediction is then combined with noisy sensor measurements to generate a maximum a posteriori (MAP) estimate of the staircase over time, effectively filtering out noise and accounting for occlusions. This MAP estimate is fused with 3D point cloud data to enable accurate segmentation of the stair surfaces, differentiating them from surrounding clutter. The contributions of this letter are:

\begin{itemize}
   \item \textbf{Staircase Modeling:} A novel split state-space model to parameterize and represent large-scale staircases
    \item \textbf{MAP Estimation:} A robust pipeline for estimating staircase location and parameters over time using noisy detections in scenes with occlusion and clutter
    \item \textbf{Stair Surface Segmentation:} An algorithm combining the staircase estimate with the point cloud to segment the staircase ground points in presence of clutter
\end{itemize}
We implement our proposed method both on real robots and in simulation in diverse environments with large and cluttered staircases (Fig.~\ref{fig:opening_results}). We achieve real-time performance with 67-89\% reduction in estimated staircase parameter errors and 30\% reduction in staircase location error compared to baselines

\section{Related Work}
\subsection{Staircase Detection and Estimation}
Researchers have explored various methods for staircase detection using both image-based and point cloud-based approaches. While image-based methods~\cite{Murakami}~\cite{ilyas2023staircase} offer computational speed, they are susceptible to environmental factors like lighting conditions, and face challenges in accurately estimating 3D geometry. Point cloud methods leverage the inherent planar geometry of stairs to segment planes and estimate staircase location and parameters. Although RANSAC-based approaches are prevalent~\cite{sanchez2021staircase}~\cite{fourre2020autonomous}, they can be non-deterministic and computationally expensive. Perez-Yus et al. used a wearable RGB-D camera to detect staircases~\cite{perez2017stairs}. Westfechtel et al. \cite{westfechtel2018robust} achieved high accuracy in staircase parameter estimation using a graph-based search strategy, but required significant processing time. Qing et al.~\cite{qing2024onboard} introduced a measurement correction technique based on estimated stair plane poses, but their approach is sensitive to plane segmentation errors that can be caused by clutter on stairs. 

The authors' prior work in \cite{sriganesh2023fast} presented a fast staircase detection method and introduced a simple matching algorithm to combine multiple detections over time, which averaged individual matched stairs based on proximity. However, this approach proved susceptible to noise, leading to estimation failures due to erroneous matches. Furthermore, the absence of explicit staircase modeling and the inability to leverage existing staircase information limited its capacity to handle incomplete detections arising from the sensor's restricted field of view or occlusions.

\subsection{Ground Segmentation}
Extensive research has investigated ground segmentation techniques due to their critical role in ensuring safe robot navigation~\cite{gomes2023survey}. Elevation maps are a simple and popular approach, which transform 3D point clouds into 2.5D grid representations for ground point classification \cite{elevation_map1}. Plane fitting techniques, frequently employing RANSAC, classify points based on their proximity to a fitted plane \cite{plane_fitting1}. These methods encounter difficulties in scenarios with multiple ground planes, such as staircases, as they necessitate multiple plane estimations, which is computationally expensive.  While variations like concentric zone divisions \cite{lim2021patchwork} strive to accelerate this process, the inherent complexity of cluttered staircases persists as a challenge.

More sophisticated methods, including Markov Random Fields (MRFs) \cite{huang2021fastmrf} and Gaussian processes \cite{chen2014gaussian}, offer enhanced accuracy but demand substantial computational resources, posing challenges for mobile robots with limited compute. Cloth simulation filters \cite{zhang2016csf} present an alternative, yet their performance on stairs is highly dependent on parameter tuning. Furthermore, purely geometric methods struggle to differentiate between small, flat obstacles on stairs and adjacent steps, raising safety concerns. Learning-based methods \cite{milioto2019rangenet++}\cite{he2022sectorgsnet} integrate semantic information to improve performance, but their effectiveness remains contingent on the specific scenarios encountered during training.

\section{Staircase Modeling}

\subsection{Staircase State-Space}

\begin{figure*}[!t]
    \centering
    \begin{subfigure}[t]{.32\linewidth}
        \centering
        \includegraphics[width = \linewidth]{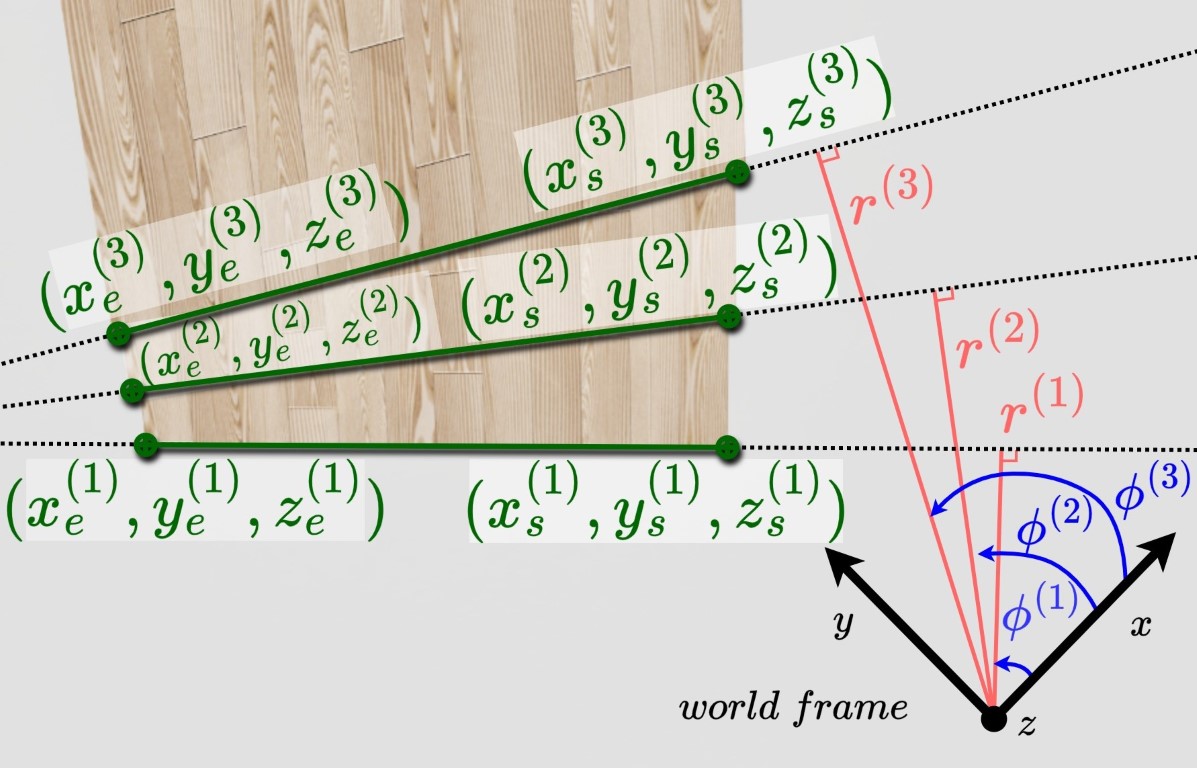}
        \caption{Top down view of the staircase.}
        \label{fig:state:top_down}
    \end{subfigure}
    \begin{subfigure}[t]{.318\linewidth}
        \centering
         \includegraphics[width = \linewidth]{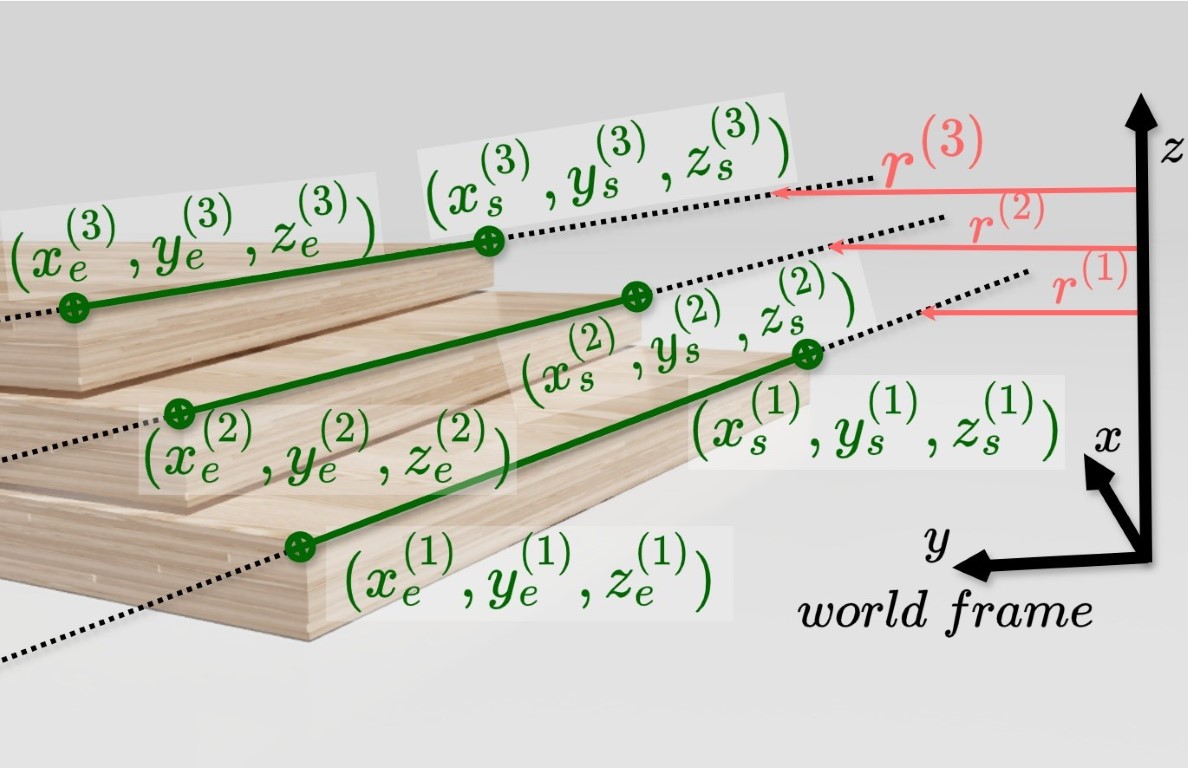}
          \caption{3-D view of the staircase.}
         \label{fig:state:3dview}
    \end{subfigure}
    \begin{subfigure}[t]{.31\linewidth}
        \centering
         \includegraphics[width = \linewidth]{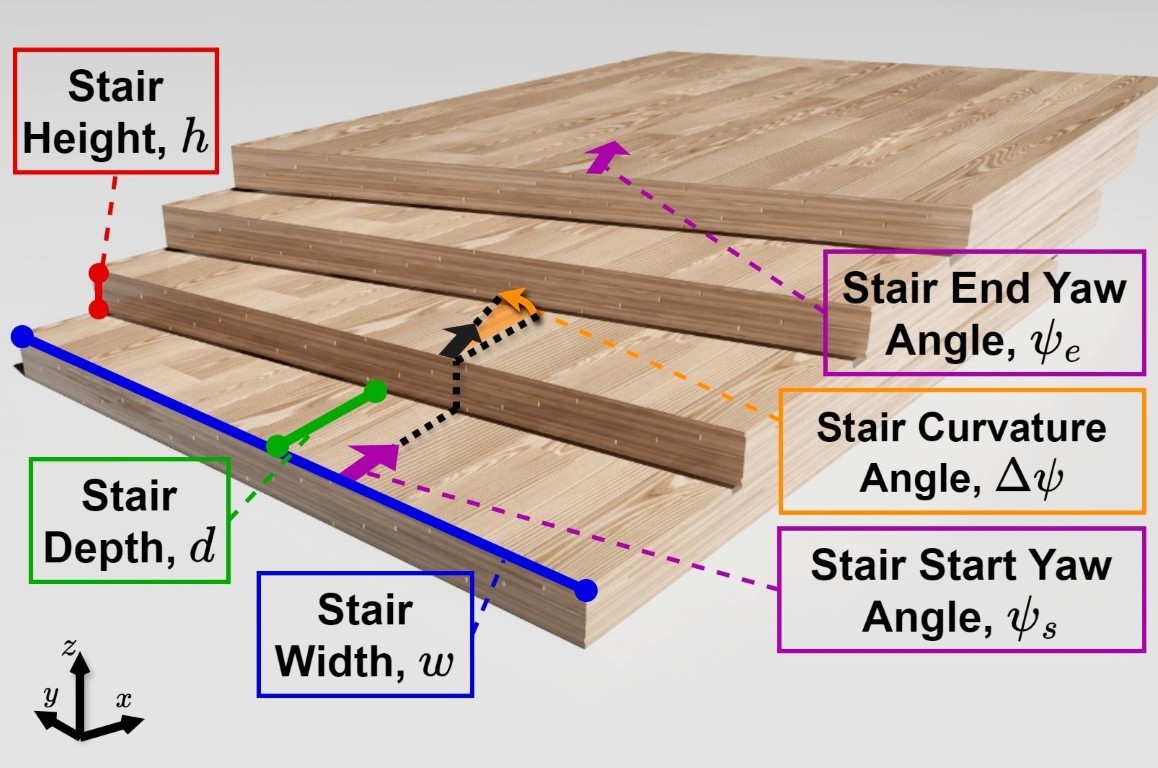}
          \caption{Model parameters for a staircase.}
         \label{fig:model_diag}
    \end{subfigure}
    \caption{Staircase annotated with the `infinite-line' and `staircase endpoint' state variables along with its model parameters.}
    \label{fig:state_fig}
    \vspace{-1.5em}
\end{figure*}
We present a split state-space model for representing staircases, enabling a robot to capture and reason about those extending beyond its sensor range. The first component, the `infinite-line staircase state' ($\mathbf{_LX}$) , represents the intersection of each stair's tread and riser as an \textit{infinite line} using polar coordinates. Subsequently, for each stair $i$ in the staircase, its state $\boldsymbol{\xi}^{(i)}$ consists of radius $r^{(i)}$, its angle $\phi^{(i)}$, along with the start and end z-positions $(z_s^{(i)}, z_e^{(i)})$ of each stair, all with respect to the robot's world frame. The infinite-line state is illustrated in Fig. \ref{fig:state:top_down}. For a staircase with $N$ steps at time step $k$, the infinite-line state vector, $\mathbf{_LX}_k$, consists of each stair line in ascending staircase direction.
\begin{align}
    \vectorbold{_LX}_k = \begin{bmatrix} \boldsymbol{\xi}^{(1)}_k \; \dots \;  \boldsymbol{\xi}^{(N)}_k \end{bmatrix}, 
    \boldsymbol{\xi}^{(i)}_k = \begin{array}{c} [r^{(i)} \; \; \phi^{(i)} \;\; z^{(i)}_s \;\; z_e^{(i)}] \end{array} \label{eq:state_vec_inf}
\end{align}

The second component, the `staircase endpoint state'~($\mathbf{_PX}$), grounds this representation by incorporating the physical location of each step. $\mathbf{_PX}$ uses cartesian coordinates to define the actual starting ($\vectorbold{p}_s$) and ending points ($\vectorbold{p}_e$) of each stair line with respect to the robot's world frame (Fig.~\ref{fig:state_fig}). This precise 3D information allows for accurate reconstruction of parts of the staircase that have been observed, while the infinite-line state can help describe potentially unseen parts of the staircase. For a staircase with $N$ steps at time step $k$, the endpoint state vector, $\mathbf{_PX}_k$, consists of each stair's endpoints in ascending staircase direction.
\begin{align}
    \vectorbold{_PX}_k &= \begin{bmatrix} \vectorbold{p}_s^{(1)} &  \vectorbold{p}_e^{(1)} & \dots & \vectorbold{p}_s^{(N)} &  \vectorbold{p}_e^{(N)} \end{bmatrix}, \label{eq:state_vec_endp} \\
    \vectorbold{p}_s^{(i)}&= \begin{array}{c} [\; x_s^{(i)} \; \; y_s^{(i)} \;\; z^{(i)}_s \;]\end{array}, \;\;\;  \vectorbold{p}_e^{(i)} = \begin{array}{c} [\; x_e^{(i)} \; \; y_e^{(i)} \;\; z_e^{(i)} \; ] \end{array} \nonumber
\end{align}
As our model represents stairs as edges (lines) it excludes the information of the initial ground plane for ascending staircases. Similarly, for descending staircases, it excludes the final ground plane at the bottom.
\subsection{Parameterized Staircase Model}
We model a staircase using six key parameters $\vectorbold{S}_k$. We assume that the parameters remain consistent between any two consecutive steps along the staircase. As depicted in  Fig.~\ref{fig:model_diag}, the six parameters of our staircase model are:
\begin{itemize}
    \item \textit{Stair Height ($h$):} The vertical distance between consecutive steps along the entire staircase.
    \item \textit{Stair Depth ($d$):} The horizontal depth of each step. In the case of curved staircases, the depth is measured from the center of each stair line, as illustrated in Fig. \ref{fig:model_diag}. This ensures consistent measurement even when the stair edges are not parallel.
    \item \textit{Stair Width ($w$):} The overall width of the staircase.
    \item \textit{Stair Start Yaw Angle ($\psi_s$):} The orientation of the first stair line with respect to the world frame, defined as the angle perpendicular to the stair line in ascending direction of the staircase. 
    \item \textit{Stair End Yaw Angle ($\psi_e$):} The orientation of the last stair line with respect to the world frame. This is defined similar to the stair start yaw angle. 
    \item \textit{Stair Curvature Angle ($\Delta\psi$):} The change in angle between consecutive steps, which captures the curvature of the staircase.
\end{itemize}
Each of these parameters can be computed directly from the current staircase state $(\vectorbold{_{L}X}_k ,\vectorbold{_{P}X}_k)$. To minimize the impact of noise, the stair height, depth, width, and curvature angle of the staircase are determined by averaging the corresponding values across all individual steps. Staircases divided by a landing are treated as multiple, separate staircases. The pronounced difference in stair depth at a landing is used to detect these instances. It's important to distinguish the stair start and end angles $(\psi_s, \psi_e)$ from the infinite-line state angles $(\phi^{(1)}, \phi^{(N)})$. While $\psi_s$ and $\psi_e$ represent the orientation of the perpendicular to the stair line in the ascending direction of the staircase, $\phi^{(i)}$ simply represents the angle of the normal to the line. Hence, based on the staircase's location in the map $\psi_s$ can be the same as $\phi^{(1)}$ or $\phi^{(1)} + \pi$. The same applies to $\psi_e$ as well. 

 This parameterized model is crucial for robustly estimating individual stair locations, as it allows us to predict the location of a stair using information from its immediate neighbors. $\boldsymbol{\xi}^{(i,j)}$ denotes the state of stair $i$ predicted given the state of its \textit{adjacent stair }$j$ using the stair parameters, as shown in \eqref{eq:predict_line}. These equations incorporate several parameters to generalize for diverse staircase configurations. The parameter $\eta$ controls the direction of prediction, where $\eta = 1$ for predicting a succeeding stair ($j = i - 1$) and $\eta = -1$ for a preceding stair $(j = i + 1)$. As detailed before, the orientation of stair $j$ (given by $\psi_s \pm  j\Delta \psi$) can differ from the actual angle of the stair line ($\phi^{(j)}$) by $\pi$, depending on the staircase's position and the world frame's orientation. To account for this, the parameter $\rho$ is introduced. Moreover, since a curved staircase has different depths at different points on the line, the parameters $\gamma$ and $l$ account for this difference and correct the stair depth ($d$) based on distance from the center of the given stair ($c_x^{(j)}, c_y^{(j)}$). 
\begin{align}
    &\begin{array}{cc}
        r^{(i,j)} = r^{(j)} + \eta \rho (d \; - l \sin (\gamma (\eta \Delta \psi)) &  z_s^{(i,j)} = z_s^{(j)} + \eta h  \\
      \phi^{(i,j)}= \phi^{(j)} + \eta(\Delta \psi) &  z_e^{(i,j)} = z_e^{(j)} + \eta h 
    \end{array} \nonumber \\
    & \eta = \begin{cases} 1 & \text{if } j = i - 1 \text{ (predicting next stair)}, \\ -1 & \text{if } j = i + 1 \text{ (predicting previous stair)}\end{cases} \nonumber \\
    & \rho = \begin{cases} 1 & \text{if } (\psi_s + j\eta\Delta\psi - \phi^{(j)}) \approx 0, \\
                          -1 & \text{if } (\psi_s + j\eta\Delta\psi - \phi^{(j)}) \approx \pi \end{cases} \nonumber \\
    & \gamma = \begin{cases} 1 & \text{if } \phi^{(i,j)} \in [\phi^{(i)},  \tan^{-1}(c_y^{(j)}, c_x^{(j)}) ], \\ -1 & \textit{otherwise} \end{cases} \nonumber \\
    & l = \sqrt{   (c^{(j)}_x - r^{(j)}\cos(\phi^{(i,j)}))^2 + (c^{(j)}_y - \;r^{(j)}\sin(\phi^{(i,j)}))^2}, \nonumber \\
    & c^{(j)}_x = (x_s^{(j)} + x_e^{(j)})/2 ,\; \; \; \; \;  c^{(j)}_y = (y_s^{(j)} + y_e^{(j)})/2  \label{eq:predict_line}
\end{align}
Similarly, we can predict the endpoint state for stair $i$ given neighboring stair $j$, denoted as $\vectorbold{p}_{s}^{(i,j)}$ and $\vectorbold{p}_{e}^{(i,j)}$ as shown in \eqref{eq:predict_endp}. The prediction starts from the known endpoint of stair $j$ and extrapolates them based on the direction. The predictions conform to the direction of extrapolation, as it utilizes the starting stair yaw angle ($\psi_s$) when predicting a succeeding stair and the final stair yaw angle ($\psi_e$) when predicting a preceding stair. Furthermore, the prediction incorporates variations in step depth  ($d$) for curved staircases by accounting for the curvature ($\Delta \psi$). 
\begin{align}
 & x_s^{(i,j)} = x_s^{(j)} + d_s \cos(\psi_o + \eta (j + 1)\Delta \psi) \nonumber \\
 & y_s^{(i,j)} = y_s^{(j)} + d_s \sin(\psi_o + \eta (j + 1)\Delta \psi) \nonumber \\
 & x_e^{(i,j)} = x_e^{(j)} + d_e \cos(\psi_o + \eta (j + 1)\Delta \psi) \nonumber \\
 & y_e^{(i,j)} = y_e^{(j)} + d_e \sin(\psi_o + \eta (j + 1)\Delta \psi) \nonumber \\
 & d_s = d + \frac{w}{2}\:\sin(\Delta \psi), \;\;\;\; d_e = d - \frac{w}{2}\:\sin(\Delta \psi), \nonumber \\
 & \psi_o = \psi_e, \; \eta = -1 \;\;\;\textit{if } \; j = i + 1 \; \; \textit{(predict previous stair)}, \nonumber \\
  & \psi_o = \psi_s, \; \eta = 1 \;\;\;\textit{if } \; j = i - 1 \; \; \textit{(predict next stair)} \label{eq:predict_endp}
\end{align}
We can estimate the state of non-adjacent stairs (i.e., stairs where  $j$ is not an immediate neighbor of $i$) by iteratively applying \eqref{eq:predict_endp}, effectively enabling the prediction of any stair along the staircase. This is particularly useful to predict the state of multiple new stairs based on the information from a single observed stair.
\subsection{Staircase Measurements and Initialization}
We leverage raw detections from our staircase detection algorithm \cite{sriganesh2023fast} as measurements ($\vectorbold{_LZ}_k$, $\vectorbold{_PZ}_k$). Measurements only capture the portion of the staircase within the robot's current field-of-view and are expressed in the robot's local frame. They are split into two components: measured stair lines ($\vectorbold{_LZ}_k$) and measured stair endpoints ($\vectorbold{_PZ}_k$). In contrast, the staircase state $(\vectorbold{_{L}X} ,\vectorbold{_{P}X})$ provides an estimate of the entire staircase structure in the world coordinate frame. This local frame measurement approach aids in decoupling uncertainty in detection and the robot localization uncertainty. The robot's local frame is defined with its x-y plane parallel to the world x-y plane, simplifying the corresponding robot's pose ($\boldsymbol{\Omega}_k$) to its position ($t_r$) and yaw orientation ($\theta_r$).
\begin{align}
    \vectorbold{_LZ}_k &= \begin{bmatrix}
                    \boldsymbol{\zeta}_{k}^{(1)}\; \dots \; \boldsymbol{\zeta}_{k}^{(M)}
                    \end{bmatrix},
     \boldsymbol{\zeta}_{k}^{(i)} = \begin{array}{c}[r_m^{(i)} \; \phi_m^{(i)} \; z^{(i)}_{sm} \; z_{em}^{(i)}] \end{array},\label{eq:measurement_vec_line} \\
    \vectorbold{_PZ}_k &= \begin{bmatrix}
                    \vectorbold{p}_{sm}^{(1)} &  \vectorbold{p}_{em}^{(1)} & \dots & \vectorbold{p}_{sm}^{(M)} &  \vectorbold{p}_{em}^{(M)}
                    \end{bmatrix}, \label{eq:measurement_vec_endp}  \\
    \vectorbold{p}_{sm}^{(i)} &= \begin{array}{c} [\; x_{sm}^{(i)} \; \; y_{sm}^{(i)} \;\; z^{(i)}_{sm} \;]\end{array}, \;\; \vectorbold{p}_{em}^{(i)} = \begin{array}{c} [\; x_{em}^{(i)} \; \; y_{em}^{(i)} \;\; z_{em}^{(i)} \; ] \end{array}\!, \nonumber \\
    \boldsymbol{\Omega}_k &= \begin{bmatrix}
                     t_r & \theta_{r}
                    \end{bmatrix} = \begin{bmatrix}
                     x_r & y_r & z_r & \theta_{r}
                    \end{bmatrix} \label{eq:robotpose_vec}
\end{align}
To enable state estimation, we represent the infinite line state by its belief,~$bel(\boldsymbol{_LX}_{k})$ at time~$k$ with mean~$\boldsymbol{_LX}_{k}$ and the covariance~$\boldsymbol{_L\Sigma}_{k}$. Upon initial detection of a staircase, we initialize the infinite-line staircase state ($\vectorbold{_LX}_k$) by transforming the measurement to world frame. This transformation, $g(\boldsymbol{\zeta}^{(i)}_k, \boldsymbol{\Omega}_k)$ is shown in \eqref{eqn:intialize:primary}. We also initialize a state covariance matrix ($\vectorbold{_L\Sigma}_k$) for the infinite-line state as shown in \eqref{eqn:intialize:primary_covar}. This covariance incorporates both the inherent noise in the measurements ($\vectorbold{\mathcal{Q}}$) and the uncertainty in the estimated robot pose from SLAM ($\vectorbold{\Sigma}_{\boldsymbol{\Omega}}$). We model the measurement noise as zero-mean Gaussian with standard deviations $\sigma_r$, $\sigma_\phi$, $\sigma_{z_s}$, $\sigma_{z_e}$ for each component of the measured stair line ($\vectorbold{_LZ}_k$).  We also initialize the endpoint state ($\vectorbold{_PX}_k$) as shown in \eqref{eq:initialize_endp}.
\begin{align}
     \boldsymbol{\xi}^{(i)}_k &= g( \boldsymbol{\zeta}^{(i)}_k, \boldsymbol{\Omega}_k) \label{eqn:intialize:primary} \\
     \begin{bmatrix}
         r^{(i)} \\ \phi^{(i)} \\ z_s^{(i)} \\ z_e^{(i)}
     \end{bmatrix} &= \begin{bmatrix}
       r_m^{(i)} + x_r \cos(\phi_m^{(i)} + \theta_r) + y_r \sin(\phi_m^{(i)} + \theta_r )  \\ \phi_m^{(i)} + \theta_r \\ z_{sm}^{(i)} + z_r \\ z_{em}^{(i)} + z_r 
     \end{bmatrix} \nonumber \\
     \vectorbold{_L\Sigma}_k &= \vectorbold{G_z \mathcal{Q} G_z^T} + \vectorbold{G}_{\boldsymbol{\Omega}} \vectorbold{\Sigma}_{\boldsymbol{\Omega}} \vectorbold{G}_{\boldsymbol{\Omega}}^T, & \label{eqn:intialize:primary_covar} \\
    \vectorbold{G_z} &= \at{\frac{\partial g}{\partial \boldsymbol{\zeta}}}{\boldsymbol{\zeta}^{(i)}_k, \boldsymbol{\Omega}_k}  \;\; \vectorbold{G_{\boldsymbol{\Omega}}} = \at{\frac{\partial g}{\partial {\boldsymbol{\Omega}_k}}}{\boldsymbol{\zeta}^{(i)}_k, \boldsymbol{\Omega}_k}  \nonumber \\
    \vectorbold{\mathcal{Q}} &= \text{Diagonal}(\sigma_r^2, \sigma_\phi^2, \sigma_{z_s}^2, \sigma_{z_e}^2), \nonumber \\
    \vectorbold{p}_{s}^{(i)}  &= \vectorbold{R}(\theta_r) \vectorbold{p}_{sm}^{(i)} + t_r, \;\;\;\;  \vectorbold{p}_{e}^{(i)} = \vectorbold{R}(\theta_r)\vectorbold{p}_{em}^{(i)} + t_r \label{eq:initialize_endp}  \\
    &  \vectorbold{R}(\theta_r) \textit{ is the SO3 rotation matrix with pure yaw } \theta_r \nonumber
\end{align}
It is important to note that, unlike the infinite-line state, we do not maintain a covariance matrix for the stair endpoint state. This design choice is motivated by two factors. Firstly, the weighted line fitting algorithm \cite{pfister2003weighted}, employed by our staircase detection module \cite{sriganesh2023fast}, inherently handles the noise in polar coordinates during the segmentation process. Secondly, we prioritize refining the infinite-line state via the Extended Kalman Filter (EKF) (explained in Section \ref{sec:estimation}) and subsequently update the endpoints based on this refined state. 

\section{Staircase Estimation} \label{sec:estimation}
\begin{figure}[b!]
    \vspace{-1.25em}
     \centering
     \includegraphics[width = 0.75\linewidth]{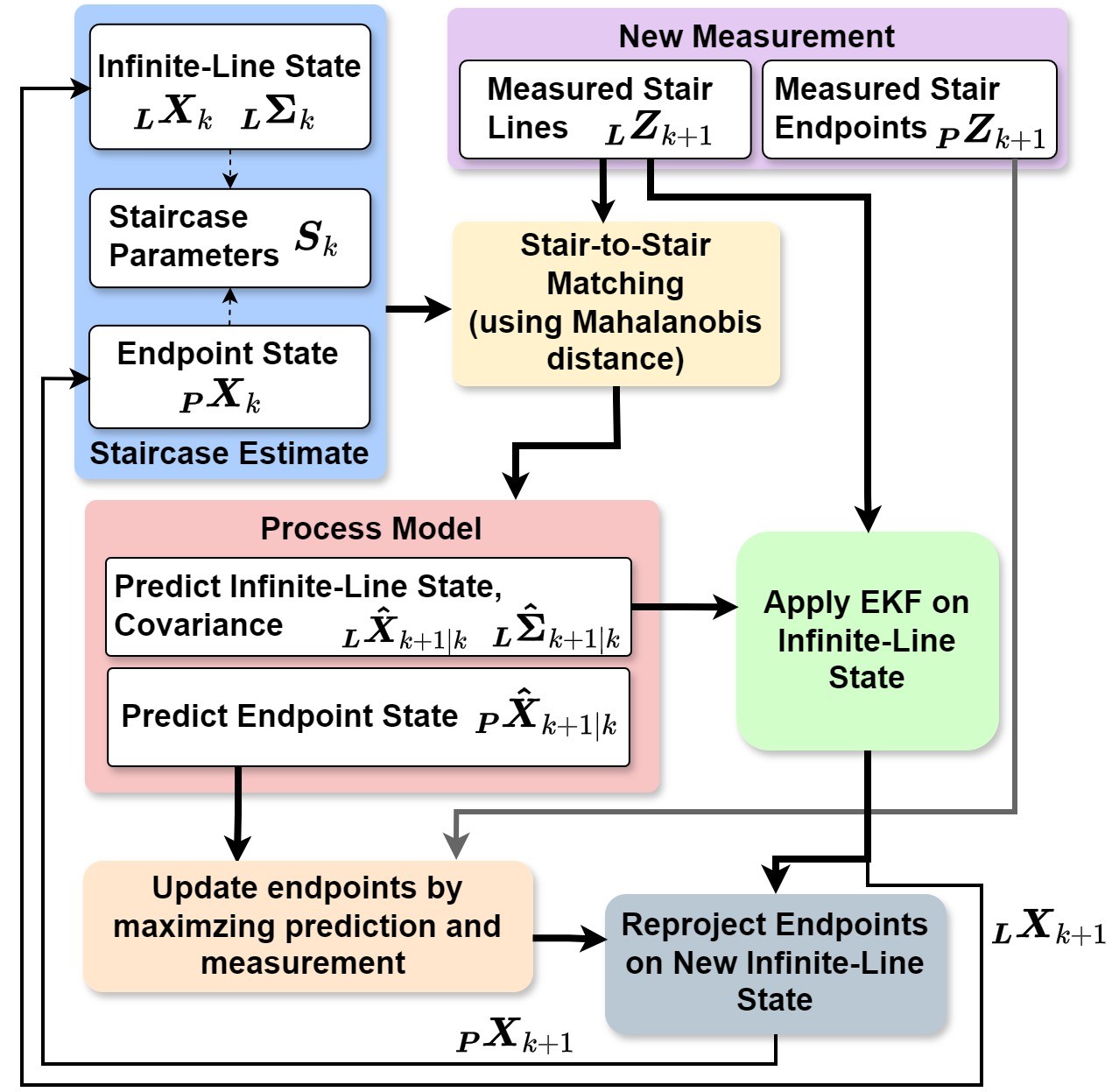}
     \caption{Block diagram depicting the overall estimation pipeline.}
     \label{fig:block_diagram}
     \vspace{-0.25em}
\end{figure}
In this section we describe our staircase estimation pipeline that utilizes our modeling framework. 
 Our pipeline operates within a Bayesian framework: we begin with a belief about the staircase state at time $k$, $bel(\boldsymbol{_LX}_{k})$. This belief is then updated at time $k+1$ to $bel(\boldsymbol{_LX}_{k+1})$ by fusing it with the incoming staircase measurement likelihood, $p(\boldsymbol{_LZ}_{k+1})$, using the Bayesian update equation:
\begin{align*}
    bel(\boldsymbol{_LX}_{k+1}) &= \beta\: p(\boldsymbol{_LZ}_{k+1})\!\! \int\displaylimits_{\boldsymbol{_LX}_{k}} \!p(\boldsymbol{_L\hat{X}}_{k+1|k}) bel(\boldsymbol{_LX}_{k}) \mathrm{d}{\boldsymbol{_LX}}
\end{align*}
Here, $\beta$ is a normalizing factor and $p(\boldsymbol{_L\hat{X}}_{k+1|k})$ represents the state transition probability obtained by the process model. Assuming Gaussian noise throughout our system, this Bayesian update could ideally be computed using a standard Kalman Filter~\cite{thrun2005probabilistic}. However, as our process model is non-linear, we employ the Extended Kalman Filter (EKF) with a linear approximation of the process model. The overall pipeline is shown in Fig. \ref{fig:block_diagram}. Further, we utilize the estimated staircase parameters to segment stair surfaces on a cluttered staircase.
\subsection{Stair-to-Stair Matching}
After initialization, new measurements undergo data association to match them with the current staircase state. This employs the Mahalanobis distance, which quantifies the similarity between a measurement and the existing state while considering uncertainty. Each stair measurement is classified as either \textit{matched}, \textit{new preceding}, or \textit{new succeeding}. Matched stairs correspond directly to existing stairs (within $3\sigma$) in the state, while new preceding/succeeding stairs indicate those located below/above the current lowest/highest stair respectively.
\subsection{Process Model - Predicting Staircase State}
The process model, denoted by  $f(\vectorbold{_L{X}}_k, \vectorbold{S}_k)$, predicts the next staircase state  $(\vectorbold{_L\hat{X}}_{k+1|k}, \vectorbold{_P\hat{X}}_{k+1|k})$ using the current state  $(\vectorbold{_L{X}}_k, \vectorbold{_P{X}}_{k})$ and staircase parameters ($\vectorbold{S}_k$). This helps track the staircase when new stairs appear or robot motion adds uncertainty. 
To accommodate new stairs, the state vector dynamically resizes, placing `new preceding stairs' at the beginning and `new succeeding stairs' at the end of the initial $N$ stairs. Thus, for $u$ new preceding and $v$ new succeeding stairs, the stairs $1$ to $u$ become the new preceding stairs, and stairs $u\!+\!N\!+\! 1$ to $u\!+\!N\! +\!v$ become the new succeeding stairs. The process model slightly differs based on this classification.

\textbf{Matched Stairs.} For stairs that matched with existing stairs in the state estimate, the process model predicts the state by incorporating information from neighboring stairs. Specifically, the predicted state of each matched stair $i$ is updated by averaging three components: its current state, the prediction from the preceding stair $(i-1)$, and the prediction from the succeeding stair $(i+1)$. For the stairs at the edges of the staircase (the lower-most and top-most existing stairs), only two predictions are averaged, as they only have one neighboring stair. Using \eqref{eq:predict_line} and \eqref{eq:predict_endp}, the predicted infinite-line state ($\boldsymbol{\hat{\xi}}^{(i)}_{k+1|k}$)  and the predicted endpoint state ($\vectorbold{\hat{p}}^{(i)}_s, \vectorbold{\hat{p}}^{(i)}_e$) for a matched stair $i$ can be computed as,
\begin{align}
 \boldsymbol{\hat{\xi}}^{(i)}_{k+1|k} &= \frac{1}{3} \left( \boldsymbol{\xi}_k^{(i)} + \boldsymbol{\xi}_k^{(i,i-1)} +  \boldsymbol{\xi}_k^{(i,i+1)} \right), \nonumber \\
\vectorbold{\hat{p}}^{(i)}_s &= \frac{1}{3} \left( \vectorbold{p}_s^{(i)} + \vectorbold{p}_s^{(i,i-1)} + \vectorbold{p}_s^{(i,i+1)} \right), \nonumber \\
\vectorbold{\hat{p}}^{(i)}_e &= \frac{1}{3} \left( \vectorbold{p}_e^{(i)} + \vectorbold{p}_e^{(i,i-1)} + \vectorbold{p}_e^{(i,i+1)} \right), \nonumber \\
\textit{where } i &\in \begin{bmatrix} u\!+\!1 & \dots & u\!+\!N \end{bmatrix}\!. \label{eq:match_predict}
\end{align}
\textbf{New Preceding/New Succeeding Stairs.} When new stairs are detected, the process model iteratively predicts their states using \eqref{eq:predict_line} and \eqref{eq:predict_endp}. For preceding stairs, we extrapolate from the lower-most existing stair (current index is $u\!+\!1$) to predict the closest preceding stair and progress downwards, and can be computed as:
\begin{align}
 & \boldsymbol{\hat{\xi}}^{(i)}_{k+1|k} =   \boldsymbol{\xi}_k^{(i ,u + 1)},  \;\;\;\;\;
\vectorbold{\hat{p}}^{(i)}_s = \vectorbold{p}_s^{(i,u+1)}, \;\;\;\;\;
\vectorbold{\hat{p}}^{(i)}_e = \vectorbold{p}_e^{(i,u+1)}, \nonumber \\
& \text{where } i \in \begin{bmatrix} u & u\!-\! 1 & \dots & 1\end{bmatrix}\!. \label{eq:preceed_predict}
\end{align}
For succeeding stairs, we extrapolate from the top-most existing stair (current index is $u\!+\!N$) to predict the closest succeeding stair and progress upwards, using:
\begin{align}
& \boldsymbol{\hat{\xi}}^{(i)}_{k+1|k} =  \boldsymbol{\xi}_k^{(i , u + N)}, \;\;\;\;\;
\vectorbold{\hat{p}}^{(i)}_s = \vectorbold{p}_s^{(i, u + N)},  \;\;\;\;\;
\vectorbold{\hat{p}}^{(i)}_e = \vectorbold{p}_e^{(i, u + N)}, \nonumber  \\
& \text{where } i \in \begin{bmatrix} u\!+\!N\!+\!1 & \dots & u\!+\!N\!+\!v \end{bmatrix}\!.  \label{eq:suceed_predict}
\end{align}
\textbf{Covariance Prediction.} The predicted covariance matrix ($\vectorbold{_L\hat{\Sigma}_{k+1|k}}$) not only reflects the uncertainty propagated from the current state but also incorporates the uncertainty associated with the staircase parameters themselves. This is achieved through the $\boldsymbol{\mathcal{R}}$ matrix, which models the uncertainty in the staircase parameters $(\sigma_h, \sigma_d, \sigma_w, \sigma_{\psi_s}, \sigma_{\psi_n}, \sigma_{\Delta\psi})$. Crucially $\boldsymbol{\mathcal{R}}$ can be user-selected, providing a mechanism to adjust the trust in the process model. Higher values in $\boldsymbol{\mathcal{R}}$ indicate less trust, causing the filter to rely more on incoming measurements. 
\begin{align}
    \vectorbold{_L\hat{\Sigma}_{k+1|k}} &= \vectorbold{F_{X}}\; \vectorbold{_L{\Sigma}_k} \; \vectorbold{F^T_{X}} + \vectorbold{F_{S}} \boldsymbol{\mathcal{R}} \vectorbold{F^T_{S}},\label{eq:predict_covar} \\
    \vectorbold{F_{X}} &= \at{\frac{\partial f}{\partial \vectorbold{_L{X}}}}{{\vectorbold{_LX}_k}, \vectorbold{S}_k} \; \begin{array}{c} \text{Jacobian of prediction} \\ \text{w.r.t infinite-line state,} \end{array} \nonumber \\
    \vectorbold{F_{_S}} &= \at{\frac{\partial f}{\partial \vectorbold{S}}}{{\vectorbold{_LX}_k}, \vectorbold{S}_k} \; \begin{array}{c} \text{Jacobian of prediction} \\ \text{w.r.t staircase parameters.} \end{array} \nonumber
\end{align}
\subsection{Updating Staircase Estimate}
The staircase state is updated in a two-step process using new measurements. First, the predicted infinite-line state~($\vectorbold{_L\hat{X}_{k+1|k}}$) is refined through an Extended Kalman Filter~(EKF). This involves transforming the predicted state to the robot's local frame and incorporating the new measurement data~($\vectorbold{_LZ_{k+1}}$) and its uncertainty ($\vectorbold{\mathcal{Q}}$). This step produces a maximum a posteriori estimate of the infinite-line staircase state $\vectorbold{_LX_{k+1}}$. If a stair has no match in the new measurement, EKF is not applied and its state remains unchanged.

Next, the endpoint state is updated through a maximization process that refines the endpoints of each stair to maximize its stair width. For each stair $i$, this process considers both the predicted endpoints from the process model ($\vectorbold{\hat{p}}^{(i)}_s, \vectorbold{\hat{p}}^{(i)}_e$) and the measured endpoints ($\vectorbold{p}^{(i)}_{sm}, \vectorbold{p}^{(i)}_{em}$), selecting the two endpoints that maximize the XY Euclidean distance. This ensures that the endpoint state is updated upon observing previously unseen stair segments. Finally, these refined endpoints are re-projected onto the updated infinite-line state ($\vectorbold{_LX_{k+1}}$) to ensure consistency between the two representations. This re-projection effectively refines the endpoint estimates by incorporating the updated stair line location obtained from EKF. The resulting staircase estimate integrates the refined infinite-line and endpoint states, providing an accurate and robust representation of the staircase structure.

\begin{figure}[b!]
    \vspace{-1.2em}
    \centering
    \begin{subfigure}[t]{0.44\linewidth}
        \centering
        \includegraphics[width = 0.95\linewidth]{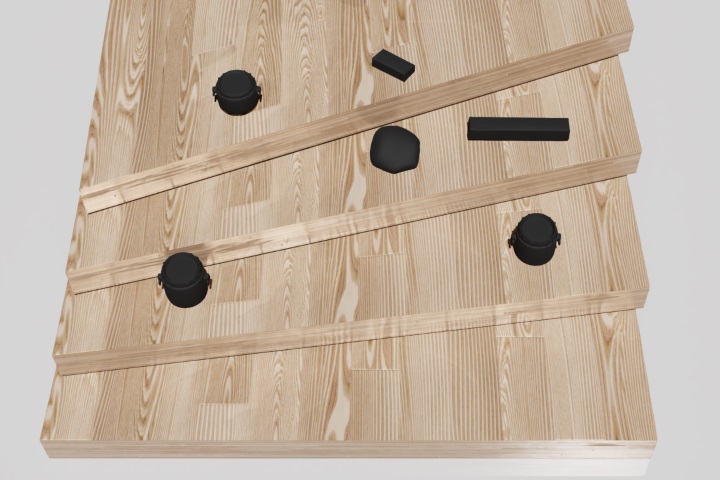}
        \caption{Cluttered Staircase Scene.}
        \label{fig:ground_plane:scene}
    \end{subfigure}
    \begin{subfigure}[t]{0.44\linewidth}
        \centering
        \includegraphics[width = 0.95\linewidth]{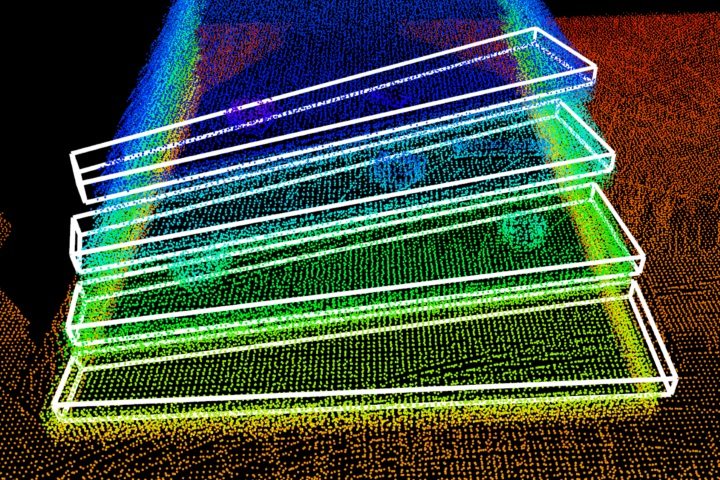}
        \caption{Generated crop-box for each stair overlaid on point cloud.}
        \label{fig:ground_plane:wireframe}
    \end{subfigure}
    \begin{subfigure}[t]{0.44 \linewidth}
        \centering
        \includegraphics[width =0.95\linewidth]{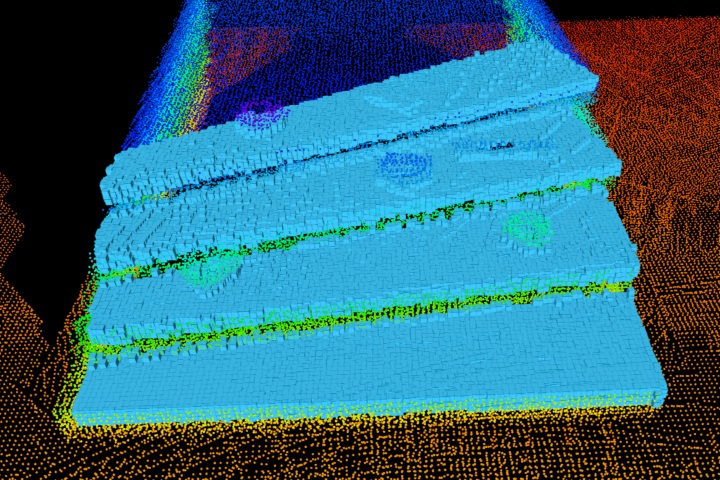}
        \caption{Points selected in the crop-box for plane fitting.}
        \label{fig:ground_plane:selected}
    \end{subfigure}
    \begin{subfigure}[t]{0.44\linewidth}
        \centering
        \includegraphics[width = 0.95\linewidth]{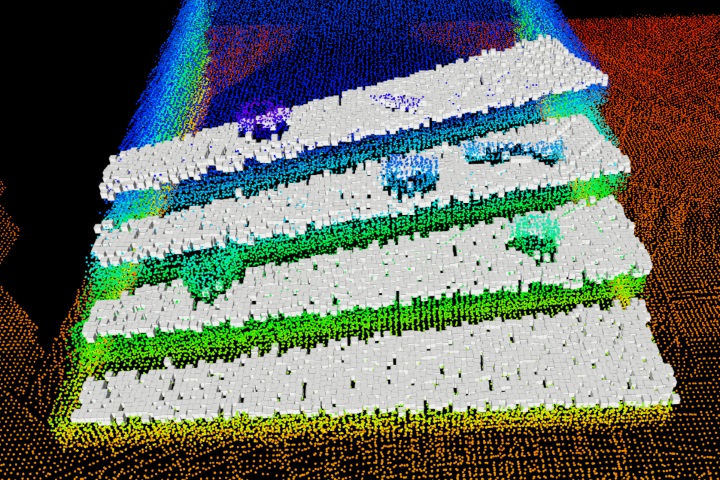}
        \caption{Final segmented stair surface points after plane-fitting.}
        \label{fig:ground_plane:final}
    \end{subfigure}
    \caption{Stair surface segmentation process for a cluttered staircase.}
    \label{fig:ground_plane}
    \vspace{-0.25em}
\end{figure}

\subsection{Stair Surface Segmentation}
Once the staircase state and the staircase parameters are updated, we can utilize this information to segment stair surfaces from cluttered scenes on each stair. For each individual stair, a 3D crop-box is defined. The length and width of the crop-box is computed using the estimated endpoints and staircase depth, whereas the height of the crop-box is derived by the uncertainty in the estimated z-positions ($z^{(i)}_s, z^{(i)}_e$) of the stair. This effectively isolates the point cloud region corresponding to each stair, as visualized in Fig. \ref{fig:ground_plane:wireframe}.

For points within each crop-box, the algorithm further refines the segmentation by fitting a plane through the points using RANSAC. Crucially, this plane fitting incorporates a constraint ensuring that the identified plane remains parallel to the ground plane. This constraint leverages the inherent geometric property of stairs, where the tread surface is typically parallel to the ground. The inliers to this fitted plane are then classified as the stair tread surface, effectively separating them from any remaining clutter or outliers within the crop-box. This approach, illustrated in Fig. \ref{fig:ground_plane}, enables robust segmentation of stair surfaces even in cluttered environments. As our staircase state excludes the initial ground plane, it is not segmented in our algorithm. Incorporating this ground plane is left as future work.

\section{Experiments and Results}
\subsection{Experimental Setup}
\subsubsection{System Overview}
Experiments were conducted both in simulation (using Isaac Sim) and real-world environments (using a Boston Dynamics Spot robot). The Spot robot was equipped with an NVIDIA Jetson AGX Orin, a Velodyne VLP-16 LiDAR for localization and mapping (LiPO~\cite{LiPo}). To enhance pointcloud density, points from the Spot's onboard RealSense sensors were registered to the LiDAR-based SLAM map. In Isaac Sim, a Velodyne VLP-128 LiDAR mounted on a simulated Spot robot generated high-fidelity point cloud data. Data collection involved manually teleoperating Spot to navigate around the staircase, ensuring complete capture by the onboard sensors.  Manual teleoperation was crucial because the challenging nature of cluttered staircases hindered Spot's autonomous navigation.

\subsubsection{Dataset}

A comprehensive dataset of 23 distinct staircase examples (17 real-world, 6 simulation) was curated to evaluate the proposed system, encompassing a diverse range of configurations (size, length, curvature, clutter). Specifically, 10 examples featured staircases without any clutter, while the remaining 13 included various types of clutter, such as large boxes obstructing staircase visibility and smaller items on the steps, representing different levels of clutter. It included staircases with widths ranging between 1 and 10 meters and step counts ranging from 4 to 20. These included both ascending and descending orientations as well as curved and open-rise staircases.

\subsubsection{Baselines}
To evaluate our approach, we compare it against two staircase estimation baselines adapted from prior state-of-the-art~\cite{sriganesh2023fast} and one segmentation baseline. The simple averaging (AVG) merging algorithm estimates the staircase state by averaging the current estimate of each stair's endpoints with corresponding new measurements. The maximizing algorithm (MAX) is similar to the averaging algorithm, except it selects the endpoints that maximize the Euclidean distance between the start and end points, effectively maximizing the estimated stair width. For segmentation, we use a Cloth Simulation Filter~\cite{zhang2016csf} as a baseline. This method simulates a cloth draped over the estimated staircase region, conforming to the underlying geometry to approximate stair surfaces. 

\subsubsection{Error Metrics}
Ground truth data for each staircase was meticulously generated through manual segmentation, involving precise annotation of staircase locations and parameters (height, depth, width, curvature), along with the removal of debris points from the point clouds. We assess staircase estimation accuracy by evaluating errors in two aspects: \textit{Staircase Parameter RMSE}, which quantifies the root-mean-square error between estimated and ground truth staircase parameters; and \textit{Stair Location RMSE}, which evaluates individual stair location accuracy using a point-to-line distance metric, calculating the distance between estimated endpoints of each stair and the corresponding ground truth stair line, along with the stair orientation error.
\subsection{Results}
\begin{figure*}[!t]
    \centering
    \begin{subfigure}[t]{.15\linewidth}
        \centering
        \includegraphics[width = \linewidth]{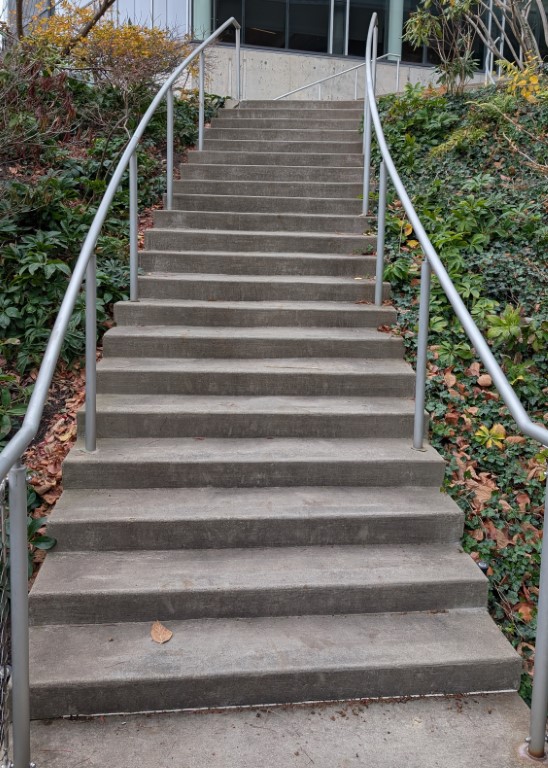}
    \end{subfigure}
    \begin{subfigure}[t]{.15\linewidth}
        \centering
         \includegraphics[width = \linewidth]{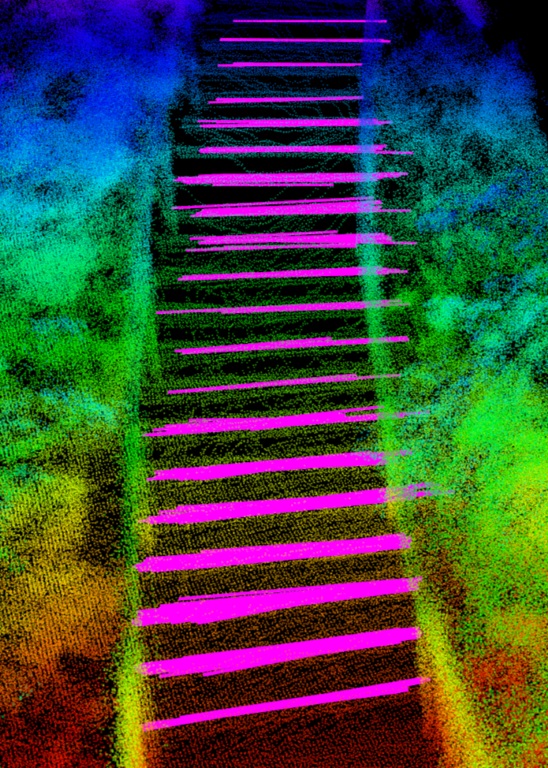}
    \end{subfigure}
    \begin{subfigure}[t]{.15\linewidth}
        \centering
         \includegraphics[width = \linewidth]{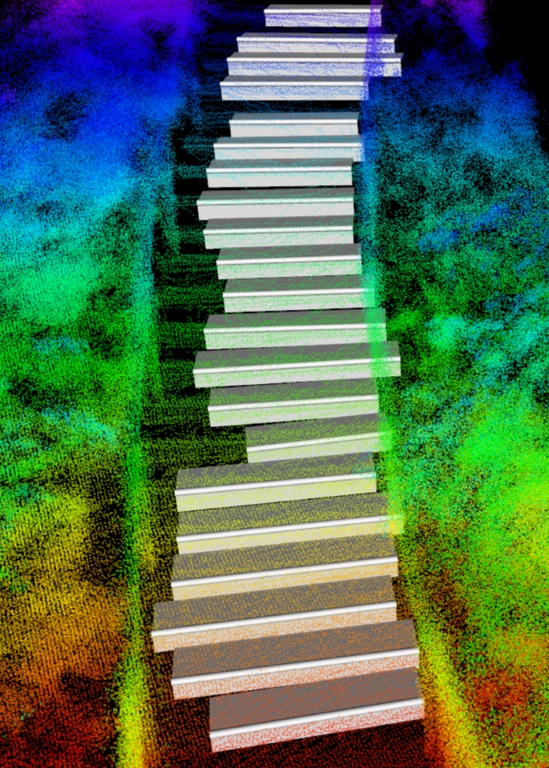}
    \end{subfigure}
    \begin{subfigure}[t]{.15\linewidth}
        \centering
         \includegraphics[width = \linewidth]{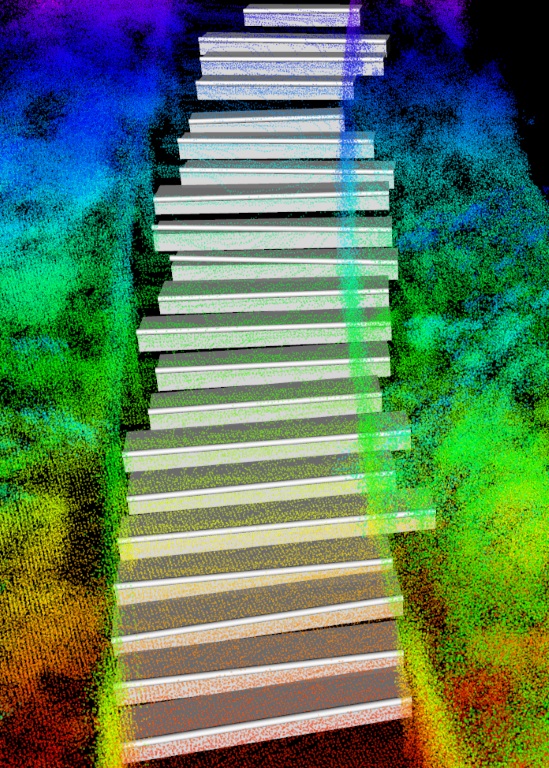}
    \end{subfigure}
     \begin{subfigure}[t]{.15\linewidth}
        \centering
         \includegraphics[width = \linewidth]{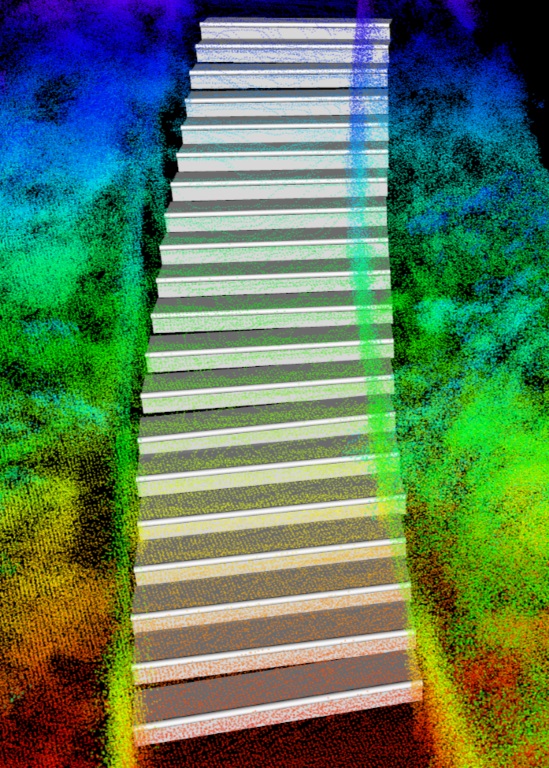}
    \end{subfigure}
    \begin{subfigure}[t]{.15\linewidth}
        \centering
        \includegraphics[width = \linewidth]{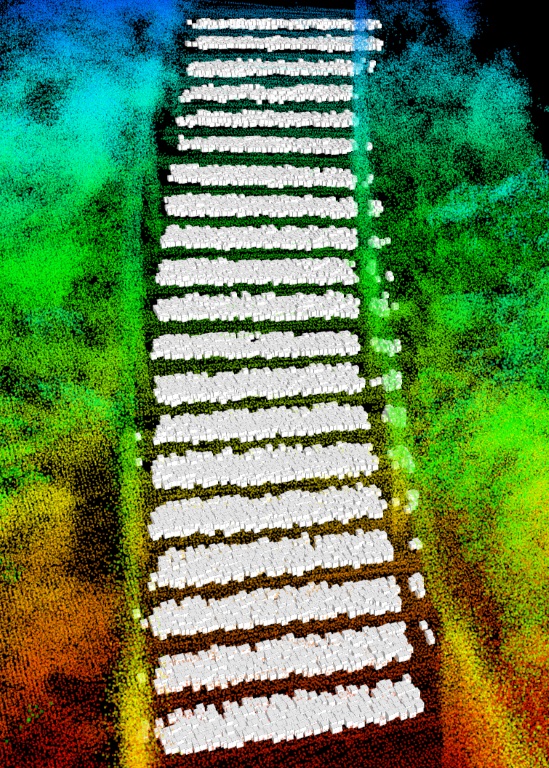}
    \end{subfigure} 
    \begin{subfigure}[t]{.15\linewidth}
        \centering
        \includegraphics[width = \linewidth]{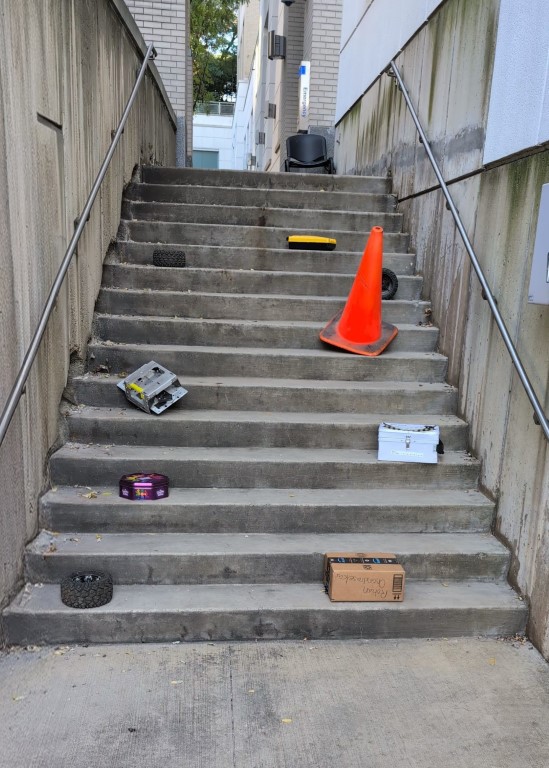}
        \caption{}
        \label{fig:results:scene}
    \end{subfigure}
    \begin{subfigure}[t]{.15\linewidth}
        \centering
         \includegraphics[width = \linewidth]{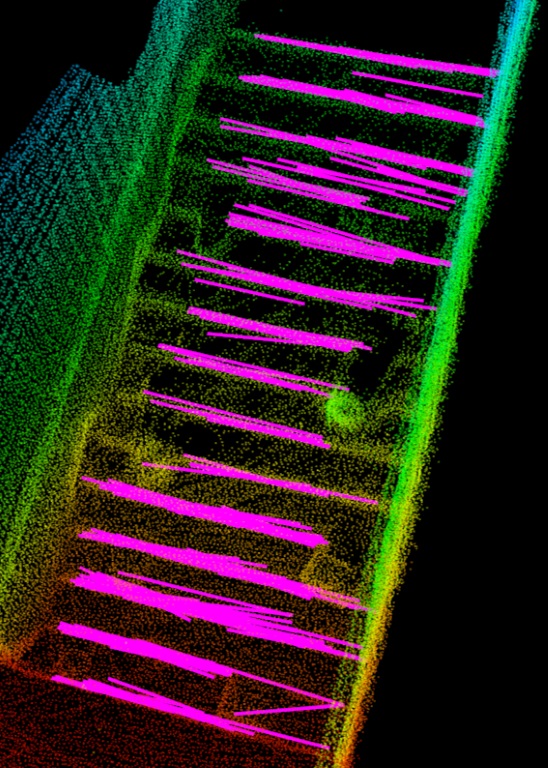}
         \caption{}
        \label{fig:results:measures}
    \end{subfigure}
    \begin{subfigure}[t]{.15\linewidth}
        \centering
         \includegraphics[width = \linewidth]{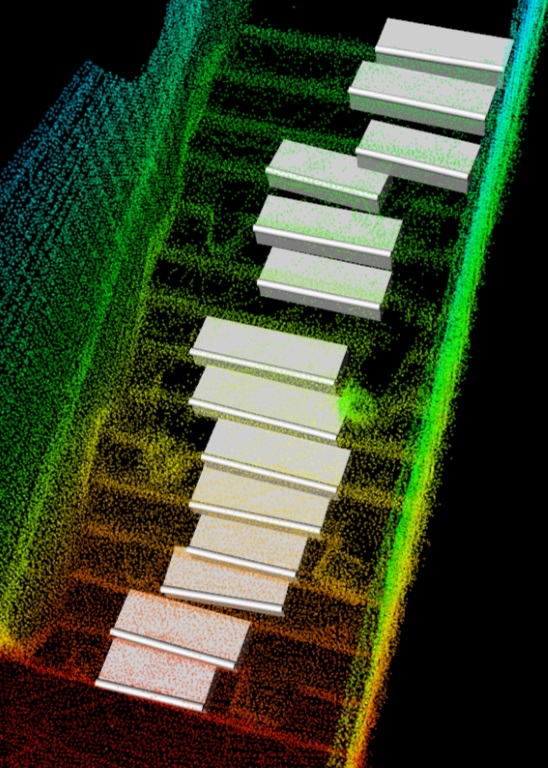}
         \caption{}
        \label{fig:results:avg}
    \end{subfigure}
    \begin{subfigure}[t]{.15\linewidth}
        \centering
         \includegraphics[width = \linewidth]{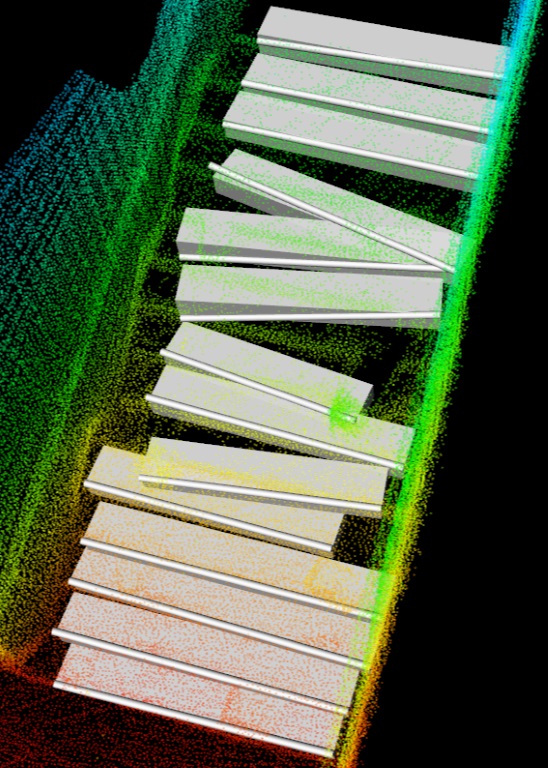}
         \caption{}
        \label{fig:results:max}
    \end{subfigure}
     \begin{subfigure}[t]{.15\linewidth}
        \centering
         \includegraphics[width = \linewidth]{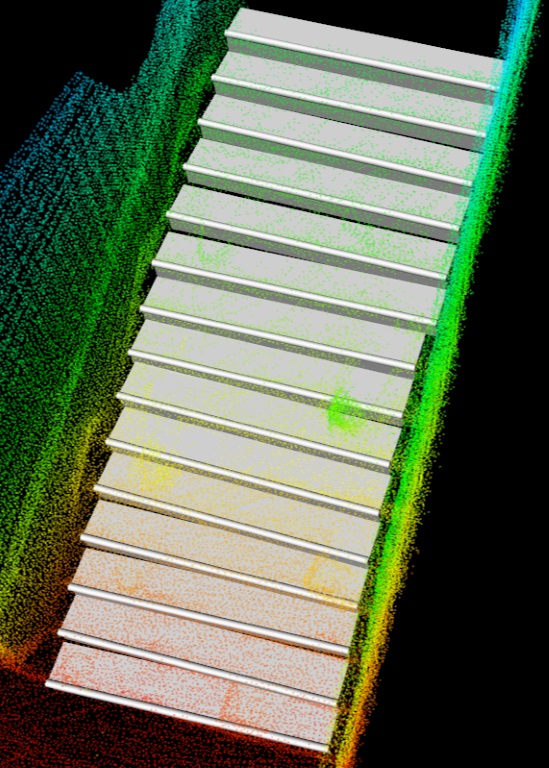}
         \caption{}
        \label{fig:results:ekf}
    \end{subfigure}
    \begin{subfigure}[t]{.15\linewidth}
        \centering
        \includegraphics[width = \linewidth]{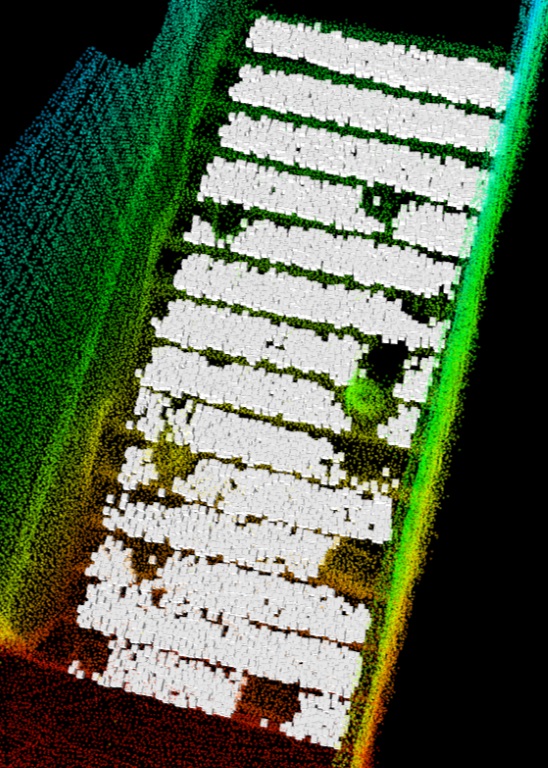}
        \caption{}
        \label{fig:results:surface}
    \end{subfigure} 
    \caption{Staircase estimation results for 2 different examples \textbf{(a)} Original staircase scenarios with curved staircase (top) and long cluttered staircase (bottom) \textbf{(b)} All input measurements (magenta) overlaid on the point cloud \textbf{(c)} Estimated staircase by using averaging algorithm~\cite{sriganesh2023fast} on the input measurements. \textbf{(d)} Estimated staircase by using maximizing algorithm on the input measurements \textbf{(e)} Estimated staircase by using our proposed framework on the input measurements \textbf{(f)} Points classified as stair surface using our segmentation algorithm}
    \label{fig:results}
    \vspace{-1.75em}
\end{figure*}

\begin{table}[b!]
\vspace{-1.5em}
\setlength{\tabcolsep}{4pt}
\centering
\begin{tabular}{ccccc}
\toprule
&  R/S & AVG \cite{sriganesh2023fast} & MAX & EKF (Ours) \\
\midrule
\multirow{2}{*}{\begin{tabular}[c]{@{}c@{}}  Stair Height\\  RMSE (cm)\end{tabular}} &  \textbf{R} & $1.8 \pm 0.8$ & $2.3 \pm 0.9$ & \color[HTML]{0f7f00} $\boldsymbol{0.6 \pm 0.3}$ \\ \cmidrule{2-5}
&  \textbf{S} & $1.1 \pm 0.5$ & $1.0 \pm 0.7$ & \color[HTML]{0f7f00}  $\boldsymbol{0.3 \pm 0.1}$ \\ \midrule
\multirow{2}{*}{\begin{tabular}[c]{@{}c@{}}  Stair Depth\\ RMSE (cm)\end{tabular}} &  \textbf{R} & $2.0 \pm 1.3$ & $3.8 \pm 2.1$ & \color[HTML]{0f7f00} $\boldsymbol{1.3 \pm 0.8}$ \\ \cmidrule{2-5}
&  \textbf{S} & $3.6 \pm 4.0$ & $3.5 \pm 0.9$ & \color[HTML]{0f7f00}  $\boldsymbol{2.0 \pm 1.5}$ \\ \midrule
\multirow{2}{*}{\begin{tabular}[c]{@{}c@{}}  Stair Width\\  RMSE (cm)\end{tabular}} &  \textbf{R} &  $113.7 \pm 112.5$ &  $29.9 \pm 18.8$ &  \color[HTML]{0f7f00} $\boldsymbol{12.0 \pm 18.7}$ \\ \cmidrule{2-5}
&  \textbf{S} &  $424.0 \pm 341.8$ &  $30.7 \pm 23.7$ &  \color[HTML]{0f7f00} $\boldsymbol{11.8 \pm 8.8}$ \\ \midrule
\multirow{2}{*}{\begin{tabular}[c]{@{}c@{}}  Stair Curvature\\  RMSE (cm)\end{tabular}} &  \textbf{R} & $2.0 \pm 1.1$ & $3.3 \pm 2.8$ & \color[HTML]{0f7f00}  $\boldsymbol{1.5 \pm 0.8}$ \\ \cmidrule{2-5}
&  \textbf{S} & $11.5 \pm 26.1$ & $1.4 \pm 1.7$ & \color[HTML]{0f7f00}  $\boldsymbol{0.8 \pm 0.7}$ \\ \midrule
\multirow{2}{*}{\begin{tabular}[c]{@{}c@{}}  Stair Location \\ XY RMSE (cm)\end{tabular}} &   \textbf{R} & $5.0 \pm 3.4$ & $6.8 \pm 5.5$ & \color[HTML]{0f7f00}  $\boldsymbol{3.6 \pm 1.5}$ \\ \cmidrule{2-5}
&  \textbf{S} & $4.0 \pm 2.5$ & $3.8 \pm 1.8$ & \color[HTML]{0f7f00}  $\boldsymbol{2.9 \pm 0.9}$ \\ \midrule
\multirow{2}{*}{\begin{tabular}[c]{@{}c@{}} Stair Location \\  Z RMSE (cm) \end{tabular}} &  \textbf{R} & $3.3 \pm 1.7$ & $3.4 \pm 2.1$ & \color[HTML]{0f7f00}  $\boldsymbol{2.3 \pm 1.2}$ \\ \cmidrule{2-5}
&  \textbf{S} & $1.4 \pm 0.7$ & $1.2 \pm 0.6$ & \color[HTML]{0f7f00}  $\boldsymbol{1.0 \pm 0.9}$ \\ \midrule
\multirow{2}{*}{\begin{tabular}[c]{@{}c@{}} Stair Orientation \\  RMSE (deg)\end{tabular}} &  \textbf{R} & $1.7 \pm 0.8$ & $2.6 \pm 1.9$ & \color[HTML]{0f7f00}  $\boldsymbol{1.5 \pm 0.7}$ \\ \cmidrule{2-5}
&  \textbf{S} & $0.9 \pm 1.0$ & $1.0 \pm 1.1$ & \color[HTML]{0f7f00}  $\boldsymbol{0.7 \pm 0.6}$ \\ \bottomrule
\end{tabular}
\caption{Comparison of staircase estimation RMSE. \textbf{R} indicates real-world scenarios, and \textbf{S} indicates simulation.}
\label{tab:rmse_estimation}
\vspace{-0.5em}
\end{table}

Figure \ref{fig:results} highlights the robustness of our proposed staircase estimation method in challenging real-world scenarios, featuring a curved staircase, and a long cluttered staircase with objects obstructing the steps. In the first example, the inferior performance of the baseline approaches can be noticed in the top portion of the stairs, where the estimated steps vary in height and fail to accurately capture the consistent rise of the actual staircase. Our approach, on the other hand, accurately captures the geometry of the curved staircase, including the consistent step heights throughout its entirety. In contrast to the curved staircase, the second scenario presents a long staircase with significant clutter that leads to noisy measurements. This noise causes significant issues for the baseline approaches: the averaging method completely misses an entire step in its estimation, while the maximizing approach produces a severely distorted and inaccurate estimate. Conversely, our method, which employs Mahalanobis distance for model matching, effectively accounts for this uncertainty, leading to a complete and accurate estimation.  Our approach is able to estimate the full extent of the staircase, even in the presence of a large occlusion (red cone) that obscures a portion of the stairs. This accurate staircase estimation enables the precise stair segmentation shown in Fig. \ref{fig:results:surface}, where our proposed algorithm classifies points belonging to the stair surfaces.

As shown in TABLE \ref{tab:rmse_estimation}, our proposed method demonstrates significantly improved accuracy in estimating both the parameters and location of staircases across all metrics in simulation and in real-world.  For real-world staircases, we achieve a 89\% reduction in width error and 67\% reduction in height error compared to the averaging approach~\cite{sriganesh2023fast}. Furthermore, our method achieves approximately a 30\% reduction in RMSE for stair location (XY and Z). While the maximizing algorithm addresses some limitations of this prior work, our approach consistently yields the most accurate results. It's worth noting that in simulation, the width error for the averaging method is considerably higher due to the presence of exceptionally wide ($10m$) staircases. Our superior performance is attributed to our method's ability to effectively incorporate information from the entire staircase structure, which helps to handle noise and incomplete measurements.

\begin{table}[b!]
\vspace{-1.8em}
\centering
    \begin{tabular}{lcc}
    \toprule
    Metric & \begin{tabular}[c]{@{}c@{}}Cloth Simulation \\ Filter (CSF) \cite{zhang2016csf} \end{tabular} & \begin{tabular}[c]{@{}c@{}}Proposed Stair \\ Segmentation Method\end{tabular} \\
    \midrule
    Accuracy (\%) & 86.36 & \color[HTML]{0f7f00}  \textbf{93.13} \\
    Precision (\%) & 91.19 & \color[HTML]{0f7f00}  \textbf{97.35} \\
    Recall (\%) & 94.43 & \color[HTML]{0f7f00} \textbf{95.56} \\
    \bottomrule
    \end{tabular}
    \caption{Comparison of Stair Surface Segmentation Metrics}
    \label{tab:segmentation}
\vspace{-0.5em}
\end{table}

TABLE \ref{tab:segmentation} presents the accuracy, precision, and recall of our proposed method for stair surface segmentation, compared against the Cloth Simulation Filter (CSF) \cite{zhang2016csf}. Our method consistently outperforms CSF across all metrics. This improvement stems from our method's ability to leverage information about the staircase structure, enabling more accurate differentiation between stair surface and flat debris. In contrast, CSF is purely geometric and often misclassifies such debris as ground. While CSF can be tuned to improve performance, we observed inconsistent results across different cluttered staircases with the same parameters. Our approach consistently provided accurate segmentation across all scenarios.

\begin{figure}[b!]
    \centering
    \vspace{-0.75em}
    \includegraphics[width = 0.85\linewidth]{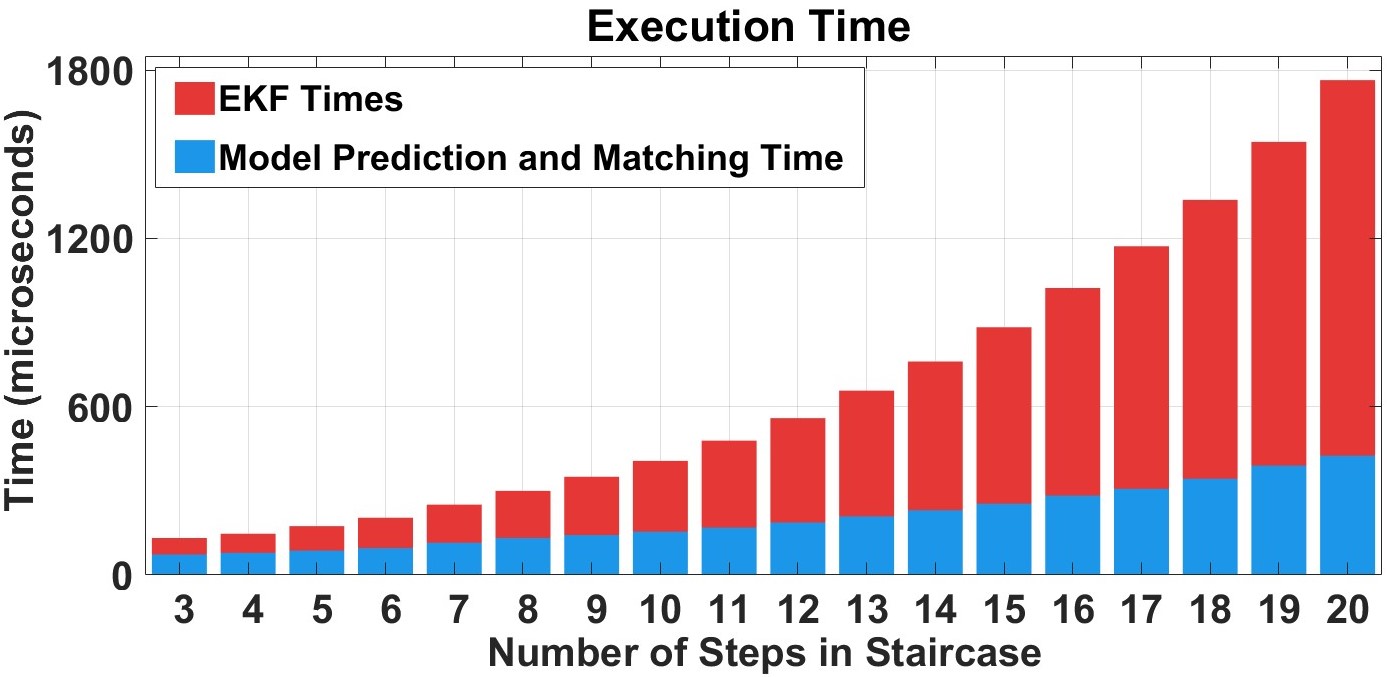}
    \caption{Plot of execution times vs number of steps for the proposed estimation pipeline}
    \label{fig:execution_time_ekf}
    \vspace{-0.9em}
\end{figure}

\begin{figure}[b!]
    \centering
    \vspace{-0.25em}
    \includegraphics[width = 0.85\linewidth]{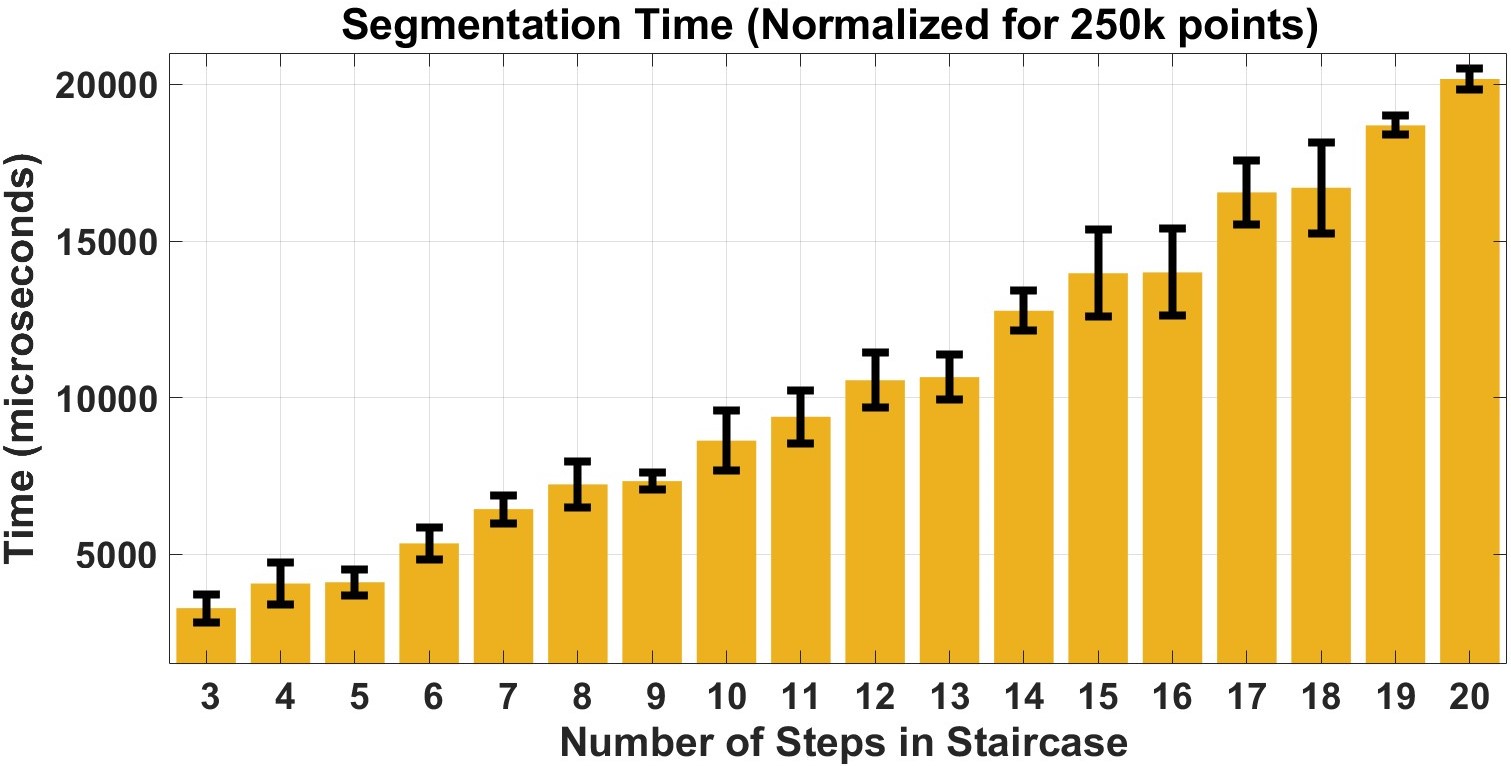}
    \caption{Plot of execution times vs number of steps for the stair surface segmentation algorithm. Error bars indicate standard deviation.}
    \label{fig:execution_time_surf}
    \vspace{-0.5em}
\end{figure}

To verify real-time performance, we evaluated our C++ implementation on an NVIDIA Jetson AGX Orin. We analyzed the computational performance of both our EKF pipeline, model prediction and matching algorithm, as well as the stair surface segmentation. Fig.~\ref{fig:execution_time_ekf} illustrates the performance of the EKF (shown in red) and the model prediction/matching (shown in blue) components, plotting the total computation time as a function of the number of stairs in the staircase (ranging from 3 to 20). Both the prediction and matching times increase with the number of stairs, with the EKF component exhibiting a steeper rise. Fig.~\ref{fig:execution_time_surf} illustrates the time required for stair surface segmentation, with times normalized for a point cloud of 250k points to account for varying point cloud densities across different staircases.  Similar to the EKF, the computation time for segmentation shows an upward trend as the number of steps increases, reflecting the increased time required for a larger number of points. Despite these trends, the absolute segmentation times remain manageable, even for staircases with 20 steps. Crucially, the total time taken by both the estimation and segmentation methods, even for large staircases, remains under 30ms. This is well within 50ms budget to achieve our real-time execution target rate of 20Hz, confirming the feasibility of real-time operation.

\section{Conclusion} 
\label{sec:conclusion}
In this paper we presented a novel approach to staircase estimation in cluttered real-world environments. We introduced a split staircase state-space model, and a Bayesian inference framework utilizing this model for robust state estimation. Our method effectively addresses challenges posed by noise, clutter, and occlusions, leading to significant improvements in accuracy compared to existing approaches.  We demonstrate the robustness and effectiveness of our approach through extensive experiments on staircases captured both on real robots and in simulation, showcasing its ability to accurately estimate staircase parameters and location even in complex scenarios. Moreover, our method achieves real-time performance, making it suitable for deployment on robotic systems operating in dynamic environments.

 While our method has proven effective for a wide range of staircases, it does have limitations due to the assumption of consistent staircase parameters. This can lead to inaccuracies when encountering irregular staircases with varying step depths or heights. To address this, we plan to explore adaptive filtering approaches to improve estimation accuracy by dynamically changing the $\vectorbold{\mathcal{R}}$ matrix based on the variations in the computed staircase parameters. We also plan to investigate the feasibility of running an ensemble of EKFs in parallel with varying $\vectorbold{\mathcal{R}}$ matrices, and selecting the estimate that minimizes the Mahalanobis distance between the residual and innovation. 
 
 In the future, we aim to utilize the segmented stair surface regions for quadruped motion planning, enabling safe navigation on cluttered staircases. Additionally, we will investigate approaches to enable fusion across multiple robots using methods such as Covariance Intersection for deployment on a multi-robot system \cite{sriganesh2024systemdesign} in multi-staircase environments.

\bibliographystyle{IEEEtran}
\bibliography{references}

\begin{thebibliography}{10}
\providecommand{\url}[1]{#1}
\csname url@rmstyle\endcsname
\providecommand{\newblock}{\relax}
\providecommand{\bibinfo}[2]{#2}
\providecommand\BIBentrySTDinterwordspacing{\spaceskip=0pt\relax}
\providecommand\BIBentryALTinterwordstretchfactor{4}
\providecommand\BIBentryALTinterwordspacing{\spaceskip=\fontdimen2\font plus
\BIBentryALTinterwordstretchfactor\fontdimen3\font minus \fontdimen4\font\relax}
\providecommand\BIBforeignlanguage[2]{{%
\expandafter\ifx\csname l@#1\endcsname\relax
\typeout{** WARNING: IEEEtran.bst: No hyphenation pattern has been}%
\typeout{** loaded for the language `#1'. Using the pattern for}%
\typeout{** the default language instead.}%
\else
\language=\csname l@#1\endcsname
\fi
#2}}

\bibitem{Murakami}
S.~Murakami, M.~Shimakawa, K.~Kivota, and T.~Kato, ``Study on stairs detection using rgb-depth images,'' in \emph{2014 Joint 7th International Conference on Soft Computing and Intelligent Systems (SCIS) and 15th International Symposium on Advanced Intelligent Systems (ISIS)}.\hskip 1em plus 0.5em minus 0.4em\relax IEEE, 2014, pp. 1186--1191.

\bibitem{ilyas2023staircase}
M.~Ilyas, A.~K. Lakshmanan, A.~V. Le, and M.~R. Elara, ``Staircase recognition and localization using convolutional neural network (cnn) for cleaning robot application,'' \emph{Mathematics}, vol.~11, no.~18, p. 3964, 2023.

\bibitem{sanchez2021staircase}
J.~A. S{\'a}nchez-Rojas, J.~A. Arias-Aguilar, H.~Takemura, and A.~E. Petrilli-Barcel{\'o}, ``Staircase detection, characterization and approach pipeline for search and rescue robots,'' \emph{Applied Sciences}, vol.~11, no.~22, p. 10736, 2021.

\bibitem{fourre2020autonomous}
J.~Fourre, V.~Vauchey, Y.~Dupuis, and X.~Savatier, ``Autonomous rgbd-based industrial staircase localization from tracked robots,'' in \emph{2020 IEEE/RSJ International Conference on Intelligent Robots and Systems (IROS)}.\hskip 1em plus 0.5em minus 0.4em\relax IEEE, 2020, pp. 10\,691--10\,696.

\bibitem{perez2017stairs}
A.~Perez-Yus, D.~Guti{\'e}rrez-G{\'o}mez, G.~Lopez-Nicolas, and J.~Guerrero, ``Stairs detection with odometry-aided traversal from a wearable rgb-d camera,'' \emph{Computer Vision and Image Understanding}, vol. 154, pp. 192--205, 2017.

\bibitem{westfechtel2018robust}
T.~Westfechtel, K.~Ohno, B.~Mertsching, R.~Hamada, D.~Nickchen, S.~Kojima, and S.~Tadokoro, ``Robust stairway-detection and localization method for mobile robots using a graph-based model and competing initializations,'' \emph{The International Journal of Robotics Research}, vol.~37, no.~12, pp. 1463--1483, 2018.

\bibitem{qing2024onboard}
C.~Qing, R.~Zeng, X.~Wu, Y.~Shi, and G.~Ma, ``An onboard framework for staircases modeling based on point clouds,'' \emph{arXiv preprint arXiv:2405.01918}, 2024.

\bibitem{sriganesh2023fast}
P.~Sriganesh, N.~Bagree, B.~Vundurthy, and M.~Travers, ``Fast staircase detection and estimation using 3d point clouds with multi-detection merging for heterogeneous robots,'' in \emph{2023 IEEE International Conference on Robotics and Automation (ICRA)}.\hskip 1em plus 0.5em minus 0.4em\relax IEEE, 2023, pp. 9253--9259.

\bibitem{gomes2023survey}
T.~Gomes, D.~Matias, A.~Campos, L.~Cunha, and R.~Roriz, ``A survey on ground segmentation methods for automotive lidar sensors,'' \emph{Sensors}, vol.~23, no.~2, p. 601, 2023.

\bibitem{elevation_map1}
A.~Asvadi, P.~Peixoto, and U.~Nunes, ``Detection and tracking of moving objects using 2.5 d motion grids,'' in \emph{2015 IEEE 18th International Conference on Intelligent Transportation Systems}.\hskip 1em plus 0.5em minus 0.4em\relax IEEE, 2015, pp. 788--793.

\bibitem{plane_fitting1}
P.~Narksri, E.~Takeuchi, Y.~Ninomiya, Y.~Morales, N.~Akai, and N.~Kawaguchi, ``A slope-robust cascaded ground segmentation in 3d point cloud for autonomous vehicles,'' in \emph{2018 21st International Conference on intelligent transportation systems (ITSC)}.\hskip 1em plus 0.5em minus 0.4em\relax IEEE, 2018, pp. 497--504.

\bibitem{lim2021patchwork}
H.~Lim, M.~Oh, and H.~Myung, ``Patchwork: Concentric zone-based region-wise ground segmentation with ground likelihood estimation using a 3d lidar sensor,'' \emph{IEEE Robotics and Automation Letters}, vol.~6, no.~4, pp. 6458--6465, 2021.

\bibitem{huang2021fastmrf}
W.~Huang, H.~Liang, L.~Lin, Z.~Wang, S.~Wang, B.~Yu, and R.~Niu, ``A fast point cloud ground segmentation approach based on coarse-to-fine markov random field,'' \emph{IEEE Transactions on Intelligent Transportation Systems}, vol.~23, no.~7, pp. 7841--7854, 2021.

\bibitem{chen2014gaussian}
T.~Chen, B.~Dai, R.~Wang, and D.~Liu, ``Gaussian-process-based real-time ground segmentation for autonomous land vehicles,'' \emph{Journal of Intelligent \& Robotic Systems}, vol.~76, pp. 563--582, 2014.

\bibitem{zhang2016csf}
W.~Zhang, J.~Qi, P.~Wan, H.~Wang, D.~Xie, X.~Wang, and G.~Yan, ``An easy-to-use airborne lidar data filtering method based on cloth simulation,'' \emph{Remote sensing}, vol.~8, no.~6, p. 501, 2016.

\bibitem{milioto2019rangenet++}
A.~Milioto, I.~Vizzo, J.~Behley, and C.~Stachniss, ``Rangenet++: Fast and accurate lidar semantic segmentation,'' in \emph{2019 IEEE/RSJ international conference on intelligent robots and systems (IROS)}.\hskip 1em plus 0.5em minus 0.4em\relax IEEE, 2019, pp. 4213--4220.

\bibitem{he2022sectorgsnet}
D.~He, F.~Abid, Y.-M. Kim, and J.-H. Kim, ``Sectorgsnet: Sector learning for efficient ground segmentation of outdoor lidar point clouds,'' \emph{IEEE Access}, vol.~10, pp. 11\,938--11\,946, 2022.

\bibitem{pfister2003weighted}
S.~T. Pfister, S.~I. Roumeliotis, and J.~W. Burdick, ``Weighted line fitting algorithms for mobile robot map building and efficient data representation,'' in \emph{2003 IEEE International Conference on Robotics and Automation}, vol.~1.\hskip 1em plus 0.5em minus 0.4em\relax IEEE, 2003, pp. 1304--1311.

\bibitem{thrun2005probabilistic}
S.~Thrun, W.~Burgard, and D.~Fox, \emph{Probabilistic Robotics}.\hskip 1em plus 0.5em minus 0.4em\relax MIT Press, 2005.

\bibitem{LiPo}
D.~Mick, T.~Pool, M.~S. Nagaraju, M.~Kaess, H.~Choset, and M.~Travers, ``Lipo: Lidar inertial odometry for icp comparison,'' \emph{arXiv preprint arXiv:2410.08097}, 2024.

\bibitem{sriganesh2024systemdesign}
P.~Sriganesh, J.~Maier, A.~Johnson, B.~Shirose, R.~Chandrasekar, C.~Noren, J.~Spisak, R.~Darnley, B.~Vundurthy, and M.~Travers, ``Modular, resilient, and scalable system design approaches - lessons learned in the years after {DARPA} subterranean challenge,'' in \emph{IEEE ICRA Workshop on Field Robotics}, 2024.

\end{thebibliography}

\end{document}